\pdfoutput=1

\documentclass[11pt]{article}

\usepackage[table,xcdraw]{xcolor}
\usepackage[final]{acl}

\usepackage{times}
\usepackage{latexsym}
\usepackage{tikz}
\usepackage{makecell}
\usepackage{siunitx}

\usepackage[T1]{fontenc}

\usepackage[utf8]{inputenc}

\usepackage{microtype}
\usepackage{booktabs}
\usepackage{multirow}
\usepackage{pifont}
\usepackage{inconsolata}

\usepackage{graphicx}
\usepackage{amsmath}

\usepackage{amsfonts} 
\usepackage{empheq} 
\usepackage{mathrsfs}
\usepackage{enumitem}
\usepackage{tabularx} 
\usepackage{ragged2e} 
\usepackage{threeparttable} 
\usepackage{float}
\usepackage{siunitx}

\usepackage{tcolorbox}
\tcbuselibrary{skins}  
\usepackage{lipsum}

\definecolor{myLightGrayBg}{RGB}{245, 245, 245}  
\definecolor{myMediumGrayHeaderBg}{RGB}{220, 220, 220} 
\definecolor{myDarkGrayBorder}{RGB}{180, 180, 180}

\colorlet{currentDefinitionBodyBg}{myLightGrayBg}
\colorlet{currentDefinitionTitleBg}{myMediumGrayHeaderBg}
\colorlet{currentDefinitionBorder}{myDarkGrayBorder}
\newtcolorbox{mydefinition}[2][]{%
  skin=bicolor,                 
  colback=currentDefinitionBodyBg,    
  colframe=currentDefinitionBorder,   
  colbacktitle=currentDefinitionTitleBg, 
  fonttitle=\bfseries,
  coltitle=black,                 
  title=#2,
  breakable,
  enhanced,
  arc=2mm,                        
  boxrule=0.8pt,                  

  fuzzy shadow={0pt}{-1pt}{-0.5pt}{0.5pt}{black!20!gray!30!white}, 
  #1
}

\usepackage{amsthm}

\newcommand{\xmark}{\ding{55}}
\newcommand{\cmark}{\ding{51}}

\newtcolorbox{takeawaybox}[2][]{
    colback=blue!5!white,  
    colframe=black,        
    colbacktitle=black,    
    coltitle=white,        
    fonttitle=\bfseries,   
    title=#2,              
    arc=2mm,               
    #1
}

\title{MERIT Feedback Elicits Better Bargaining in LLM Negotiators}

\author{
  Jihwan~Oh\textsuperscript{1}\hspace{1em}
  Murad~Aghazada\textsuperscript{1}\hspace{1em}
  Yooju~Shin\textsuperscript{2}\hspace{1em}
  Se-Young~Yun\textsuperscript{1\dag}\hspace{1em}
  Taehyeon~Kim\textsuperscript{3\dag}\\
  \textsuperscript{1}KAIST AI\hspace{1em}
  \textsuperscript{2}Amazon AGI\hspace{1em}
  \textsuperscript{3}LG AI Research
}

\begin{document}
\maketitle
\vspace{-50pt}

\begin{abstract}
Bargaining is often regarded as a logical arena rather than an art or a matter of intuition, yet Large Language Models (LLMs) still struggle to navigate it due to limited strategic depth and difficulty adapting to complex human factors. Current benchmarks rarely capture this limitation. To bridge this gap, we present a utility feedback centric framework. Our contributions are: (i) \textsc{AgoraBench}, a new benchmark spanning nine challenging settings (e.g., deception, monopoly) that supports diverse strategy modeling; (ii) human-aligned, economically grounded metrics derived from utility theory. This is operationalized via agent utility, negotiation power, and acquisition ratio that implicitly measure how well the negotiation aligns with human preference and (iii) a human preference grounded dataset with learning pipeline that strengthens LLMs’ bargaining ability through both prompting and finetuning. Empirical results indicate that baseline LLM strategies often diverge from human preferences, while our mechanism substantially improves negotiation performance, yielding deeper strategic behavior and stronger opponent awareness.
\end{abstract}

\section{Introduction}
\label{introduction}

\begin{table*}[t]
\centering
\captionsetup{skip=2pt} 
\setlength{\abovecaptionskip}{2pt}
\setlength{\belowcaptionskip}{0pt}

{\renewcommand{\arraystretch}{0.90} 
\setlength{\tabcolsep}{4pt}         
\setlength{\belowrulesep}{0pt}      
\setlength{\aboverulesep}{0pt}      

\resizebox{\textwidth}{!}{%
\begin{tabular}{@{}ll|ccccc@{}}
\toprule
 & \makecell[r]{Bargaining task} & Deal or No Deal? & Craiglist & AI feedback & AmazonHistoryPrice & \textbf{\textsc{AgoraBench}} \\
Features&&\cite{lewis2017deal}&\cite{he2018decoupling}&\cite{fu2023improving}&\cite{xia2024measuring}&(Ours) \\
\midrule
\multirow{2}{*}[-0.2em]{\textbf{Dataset}}  & Economically grounded  & \xmark & \xmark & \xmark & \xmark & \textcolor{green!60!black}{\cmark} \\ \cmidrule{2-7}
& Human Preference  & \xmark & \xmark & \xmark & \xmark & \textcolor{green!60!black}{\cmark} \\ \midrule
\multirow{3}{*}[-0.4em]{\textbf{Evaluation}} & Profit metric  & \xmark & \textcolor{green!60!black}{\cmark} & \textcolor{green!60!black}{\cmark} & \textcolor{green!60!black}{\cmark} & \textcolor{green!60!black}{\cmark} \\ \cmidrule{2-7}
& Acquisition metric    & \xmark & \xmark & \xmark & \xmark & \textcolor{green!60!black}{\cmark} \\ \cmidrule{2-7}
& Human aligned metric  & \xmark & \xmark & \xmark & \xmark & \textcolor{green!60!black}{\cmark} \\ \bottomrule
\end{tabular}%
}}
\caption{Feature comparison between bargaining research. We provide economically grounded market condition and human preference dataset. Further, we devise human preference aligned metric to evaluate the language models.}
\vspace{-12pt}
\label{tab:comparison between datasets}
\end{table*}

The emergence of Large Language Models (LLMs) as autonomous agents  is shifting strategic decision-making and negotiation tasks from humans to LLM-driven systems \cite{yao2023react, schick2023toolformer, park2023generative, xia2024measuring, wang2024interpretable, shea-etal-2024-ace, bianchi2024well, abdelnabi2024cooperation, zhu2025automated, kwon2025astra, chen2025enhancing, long2025evoemo}. Recent work on bilateral negotiation \cite{deng2024llms, bianchi2024well, shea-etal-2024-ace, xia2024measuring, kwon2024llms, zhu2025automated, kwon2025astra, chen2025enhancing, long2025evoemo} showcases LLMs’ growing proficiency in multi-turn bargaining, underscoring negotiation as a central arena for strategic reasoning.
Nonetheless, two core challenges remain: (i) the scarcity of sophisticated benchmarks that capture complex, adversarial bargaining conditions, and (ii) the difficulty of evaluating strategic planning and reasoning, especially in a human-centric sense of what constitutes a \textit{good negotiator} as opponents’ behaviors evolve dynamically beyond simple profit.

(i) Existing benchmarks \citep{he2018decoupling, lewis2017deal, xia2024measuring,fu2023improving} largely emphasize simplified, single-issue settings, limiting their usefulness for addressing the weaknesses of current LLM negotiators. Prior works \cite{xia2024measuring, kwon2024llms} indicate persistent shortcomings regarding immature Theory of Mind (ToM), constrained strategic adaptability, and often shallow reasoning. These issues are compounded by datasets \citep{he2018decoupling, lewis2017deal, fu2023improving, xia2024measuring} that largely ignore common market mechanisms such as installment plans, monopolistic structures, or negative seller perception sentiment. As a result, important facets of reliable bargaining are rarely represented, creating a sizable gap in LLMs’ ability to acquire robust negotiation competencies for intricate, realistic or economically grounded contexts.

To close this gap, we present \textsc{AgoraBench}, a benchmark that captures diverse, economically motivated market regimes (\autoref{tab:comparison between datasets}). It spans diverse market regimes (e.g., monopolistic, installment-based, negative seller perception) and product types (e.g., cameras, drones), offering a comprehensive testbed for evaluating LLM performance. By incorporating diverse conditions, \textsc{AgoraBench} provides a strong simulations for probing and improving LLMs’ capacity to manage negotiation intricacies. We also release a human preference dataset derived from LLM-agent dialogues to support in-context learning and fine-tuning.

(ii) Beyond data coverage, steering LLMs toward human-like strategic bargaining requires suitable evaluation and feedback. Since LLMs may otherwise default to naive optimization, we introduce \textsc{Merit}, a new, multi-dimensional metric designed to align behavior with human preferences, moving past conventional profit-centric assessments \citep{he2018decoupling, lewis2017deal, xia2024measuring, fu2023improving}. We observe that profit-alone measures miss genuine human priorities (\S \ref{subsec:human_preference_alignment}). For example, in multi-product settings, securing the desired item can be valued more than minimizing price. By combining profit with acquisition ratio, \textsc{Merit} better captures human preferences by accounting for both the negotiation process and its outcomes. In our benchmark, higher \textsc{Merit} indicates a negotiator that behaves more consistently with human bargaining preferences, rather than optimizing numerical utility alone.

To this end, we utilize \textsc{Merit} as a feedback to promote exploration of strategic planning and reasoning against adaptive opponents. Integrating this metric into negotiation via an in-context learning (ICL) and fine-tuning enables LLMs to internalize the nuanced trade-offs of realistic bargaining. \autoref{tab:comparison between datasets} highlights the greater scenario diversity and human-aligned evaluation of our framework relative to prior work. Our contributions are as follows.
\vspace{-15pt}

\begin{itemize}
\item \textsc{AgoraBench}. A benchmark capturing a wide range of scenarios that reflect realistic market environments. (\S \ref{sec:bargain_arena})

\item \textsc{Merit}. A novel, multi-faceted evaluation metric that comprehensively assesses buyer-side bargaining performance. Grounded in economic theory, it aligns with human preferences over both process and outcomes. (\S \ref{sec:merit})

\item We identify negotiation behaviors exhibited by LLM negotiators that deviate from typical human negotiating patterns. (\S \ref{sec:behaviors})

\item We show that both \textsc{Merit} guided ICL and finetuning advances LLM strategic negotiation, raising deal rates over multiple models (\S \ref{sec:ICL_UF}).
\end{itemize}
\section{Related works}

\begin{figure*}[t]

\centering
\small
\centerline{\includegraphics[width=0.92\textwidth]{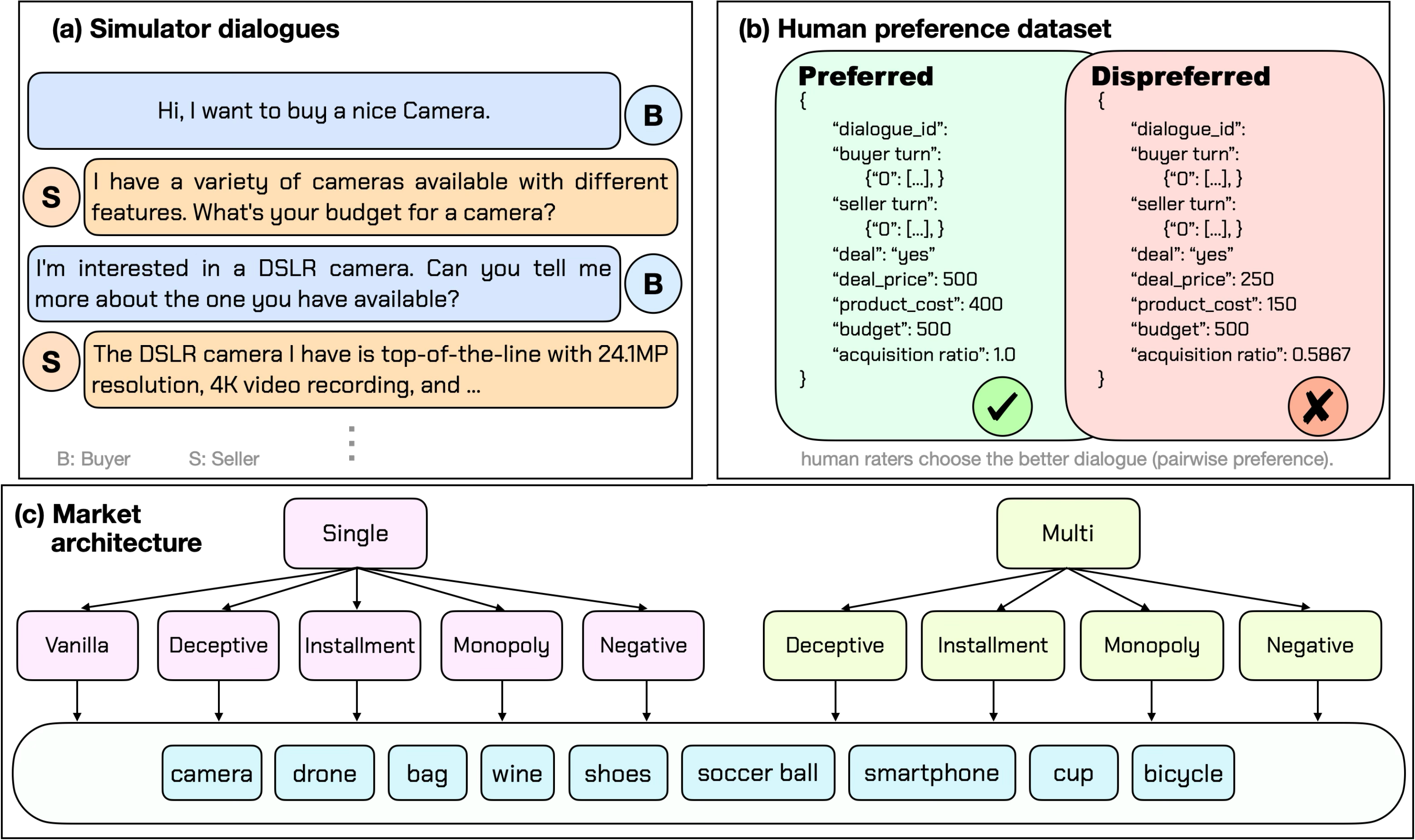}}
\vspace{-5pt}
\caption{Negotiation task between LLM agents and \textsc{AgoraBench} overview. (a) represents the simulator, (b) represents the human preference dataset, and (c) represents nine economically grounded market environments, each crafted around a distinct consumer good.}
\label{fig:main}
\vspace{-15pt}
\end{figure*}

\subsection{Negotiation and bargaining tasks}
Negotiation has drawn growing interest within natural language processing as language models increasingly act as autonomous agents capable of goal-directed, multi-turn dialogue \cite{kwon2024llms, bianchi2024well, guan2024richelieu, hua2024game, shea-etal-2024-ace, zhu2025automated, kwon2025astra}. A foundational line of bargaining work is \citet{lewis2017deal}, which introduced the \textit{Deal or No Deal?} human–human bargaining corpus and examined end-to-end models for negotiating item splits. Building on this, research shifted toward more structured strategy generation. For instance, \citet{he2018decoupling} proposed a modular approach that separates high-level strategy planning from surface realization and released the \textit{Craigslist} bargaining dataset. \citet{fu2023improving} designed a simplified balloon-product bargaining task and leveraged LLM critiques as feedback for in-context learning. \citet{xia2024measuring} constructed a large-scale bargaining dataset by harvesting Amazon product data from the camelcamelcamel platform. However, as summarized in \autoref{tab:comparison between datasets}, these resources still fall short of the diversity and real-world complexity required for robust evaluation.

\subsection{Model alignment with human preferences}
As language models increasingly operate as autonomous decision-makers with strategic reasoning capabilities, aligning their behavior with human preferences has become a central research objective. In negotiation and bargaining, performance must be judged not only by quantitative scores but also by subtler, human-centric priorities—such as fairness, perceived satisfaction, and trustworthiness \cite{weidinger2021ethical, bai2022training, perez2023discovering}. 

A substantial body of work on reward alignment seeks reward designs that more faithfully capture human intent, moving beyond coarse metrics like task accuracy or profit alone. Preference-based approaches, for example, leverage human comparisons and feedback to steer reinforcement learning agents toward human values \cite{christiano2017deep, knox2022models, ouyang2022training}. Related efforts introduce multi-objective reward formulations to encode trade-offs among competing goals—e.g., short-term profit versus long-term customer satisfaction \cite{hwang2024promptable, wang2024interpretable}. Despite these advances, comparatively little attention has been paid to embedding human preferences directly into the evaluation metrics used for bargaining tasks.

\section{\textsc{AgoraBench}}
\label{sec:bargain_arena}

\begin{figure*}[t]
\centering
\small
\centerline{\includegraphics[width=1.0\textwidth]{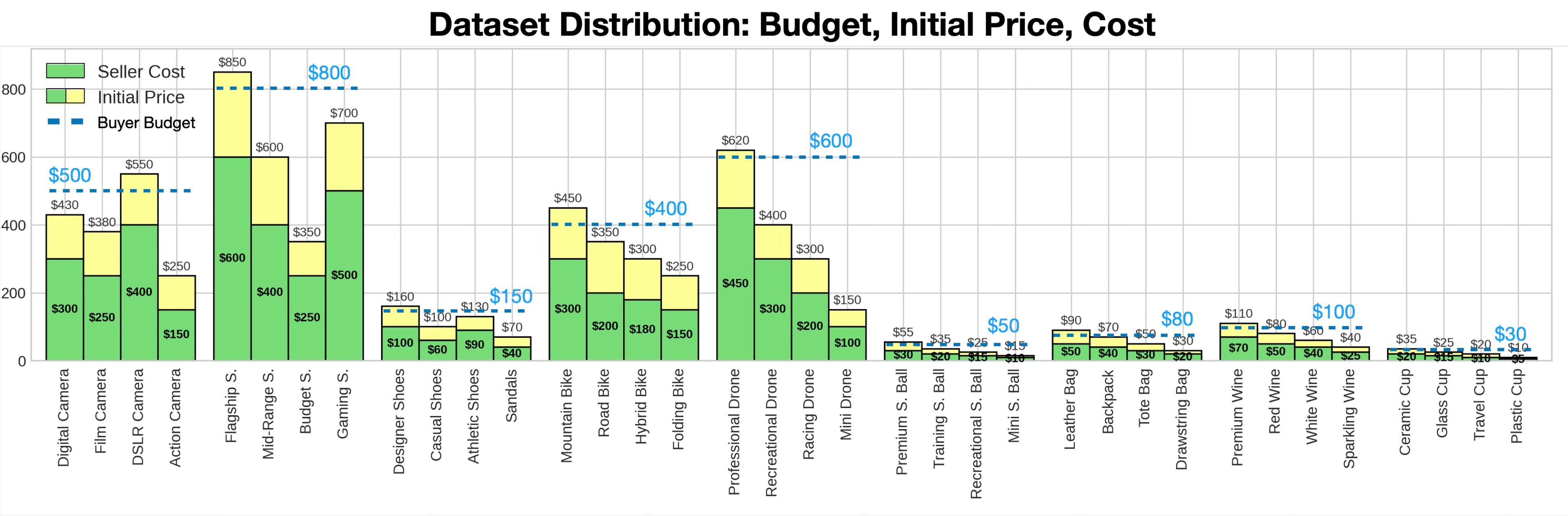}}
\vspace{-10pt}
\caption{Product attributes for negotiation scenarios in \textit{multi} product setting. We deploy seller cost, initial offer price, and buyer's budget for each product.}

\label{fig:data_distribution}
\vspace{-15pt}
\end{figure*}

\textsc{AgoraBench} mirrors real-world economic bargaining across nine distinct market types, curated to surface specific commercial negotiation challenges. It provides both an online simulator (§\ref{subsec:simulator_environment}) and a static offline dataset (§\ref{subsec:human_preference_dataset}). To ensure negotiation is necessary, we assume the seller’s initial ask always exceeds the buyer’s willingness to pay (WTP). We adopt the term \emph{WTP} over \emph{budget} to more precisely reflect economic preferences.

\subsection{Simulator environment}
\label{subsec:simulator_environment}
Two LLM agents engage in negotiations within the simulator using a structured \emph{Thoughts--Talk--Action} schema \citep{yao2023react, xia2024measuring}, also referred to as ReAct (\autoref{fig:main} (a)). Each agent operates within a constrained action space (BUY, SELL, REJECT, DEAL, and QUIT) following the framework established by \citet{xia2024measuring}. Each dialogue is grounded in a specific buyer--seller scenario defined by key hyperparameters: the market setting, agent profiles, and product attributes such as cost, willingness to pay (WTP), and initial ask. The \textsc{AgoraBench} simulator encompasses nine distinct market regimes, constructed by instantiating five market settings under the \textsc{Single} condition and four additional settings under the \textsc{Multi} condition (\autoref{fig:main} (c)). A full specification of the dataset format and fields is provided in Appendix~\ref{sec:app:data_structure}.

\begin{itemize}[itemsep=0pt, leftmargin=9pt]
\item \textbf{Vanilla market:} A baseline, product negotiation with no additional complicating factors.
\item \textbf{Deceptive market:} Settings in which either party may misrepresent information (e.g., product quality), testing agents’ ability to identify reliable signals. Except for the vanilla case, all other markets are built on this deception layer, with further conditions added atop it.
\item \textbf{Monopoly market:} A single-seller environment that limits buyer choice, enabling analysis of bargaining under asymmetric power.
\item \textbf{Installment possible market:} Introduces deferred or staggered payments, bringing time-sensitive financial trade-offs and flexible terms into the dialogue.
\item \textbf{Negative perception market:} The seller carries a prior reputational disadvantage (e.g., scandals), biasing buyers toward lower offers and creating distinct strategic dynamics as the seller attempts to counteract this effect.
\item \textbf{Single product market:} Buyers negotiate over a single item with no substitution across alternatives. This isolates bargaining behavior in a focused setting where outcomes depend solely on the agreed price and the buyer’s willingness-to-pay.
\item \textbf{Multi product market:} Buyers consider multiple items and can substitute away from a preferred option due to constraints (e.g., budget). This highlights the trade-off between product preference and cost savings. We layer this condition atop each non-vanilla market.
\end{itemize}

\subsection{Human preference}
\label{subsec:human_preference_dataset}
We further present a human preference dataset for negotiation over multiple LLM negotiations. Precisely, negotiation dialogues by multiple LLMs in deceptive multi-product negotiation settings across all set of products are collected, and then  human preferences are obtained via surveys on Amazon Mechanical Turk (MTurk)\footnote{\url{https://www.mturk.com/}}. Data is annotated in a standardized JSONL format as shown in \autoref{fig:main} (b). Using this dataset, we demonstrate how human preference signals can be leveraged to improve negotiation performance in \autoref{sec:ICL_UF}.

\section{Human-Aligned Utility Metrics}
\label{sec:merit}

We introduce \textsc{Merit}, a \underline{M}ulti-dimensional \underline{E}valuation of \underline{R}easoning \& \underline{I}nteraction in \underline{T}rade, grounded in the economic principles of \emph{utility} and \emph{bargaining}. In standard microeconomics \citep{mankiw2007principles, jehle2011advanced}, rational agents seek to maximize \emph{utility}, a construct that encompasses more than pure profit. In line with this perspective, \textsc{Merit} evaluates not only monetary gains but also an agent’s overall satisfaction from the exchange (total utility) and the degree to which the chosen product aligns with prior \emph{preferences} (ordinal utility). Thus, \textsc{Merit} captures both the \emph{cardinal} and \emph{ordinal} facets of utility that are essential for realistic evaluation of bargaining tasks.

\subsection{\textsc{Merit}: Formulation}
The performance of a language model as a buyer is quantified by \textsc{Merit} through a score reflecting how effectively the agent achieves its objectives within a given negotiation scenario. In microeconomic terms, this score corresponds to the \textit{overall satisfaction} or \textit{total benefit} an agent derives from a transaction, paralleling the notion of \textit{cardinal utility} where a measurable value is assigned to outcomes. Specifically, the \textsc{Merit} score for a buyer is a weighted combination of three key components: \textit{Consumer Surplus (CS)}, \textit{Negotiation Power (NP)}, and \textit{Acquisition Ratio (AR)}. Each component captures a distinct economic concept related to consumer behavior, bargaining dynamics, and preference fulfillment, as defined in \autoref{eq:profit_equation}:

{%
\setlength{\abovedisplayskip}{5pt plus 2pt minus 3pt}
\setlength{\abovedisplayshortskip}{0pt plus 2pt}
\begin{equation}
\vspace{-5pt}
\textsc{Merit}_{\text{buyer}} = \alpha \times \text{CS} + \beta \times \text{NP} + \gamma \times \text{AR}
\label{eq:profit_equation}
\end{equation}%
}

\paragraph{Consumer surplus}

In economics, \textit{consumer surplus} refers to the net benefit (or satisfaction) a consumer derives from purchasing a product or service \citep{mankiw2007principles}. It traditionally measures the difference between the consumer’s \textit{willingness to pay} ($P_{\text{wtp}})$ and the actual price paid ($P_{\text{deal}}$). Within \textsc{Merit}, we normalize this surplus relative to the potential surplus range determined by the seller's production cost ($P_{\text{cost}}$):
\vspace{-2pt}
\begin{equation}
\text{Consumer Surplus (CS)} = \frac{P_{\text{wtp}}-P_{\text{deal}}}{P_{\text{wtp}}-P_{\text{cost}}}
\label{eqs:CS}
\end{equation}
Bounding CS between 0 and 1 translates consumer surplus into a scaled, \textit{cardinal measure} of how effectively the buyer maximized net benefits. A value approaching 1 indicates a highly favorable deal, while 0 suggests an unfavorable outcome or an infeasible price.

\renewcommand{\arraystretch}{0.93} 
\begin{table*}[t]
\centering
\small
\setlength{\tabcolsep}{3.5pt}
\resizebox{\textwidth}{!}{
\begin{tabular}{lcccccccccc}
\toprule
\multirow{3.5}{*}{\textbf{Models}} & \textbf{Vanilla} & \multicolumn{2}{c}{\textbf{Deceptive}} & \multicolumn{2}{c}{\textbf{Monopoly}} & \multicolumn{2}{c}{\textbf{Installment}} & \multicolumn{2}{c}{\textbf{Negative Perc.}} & \multirow{3.5}{*}{\textbf{Avg}} \\
\cmidrule(lr){2-2} \cmidrule(lr){3-4} \cmidrule(lr){5-6} \cmidrule(lr){7-8} \cmidrule(lr){9-10}
 & Single & Single & Multi & Single & Multi & Single & Multi & Single & Multi & \\
\midrule
\multicolumn{11}{l}{\textit{\textbf{GPT-series}}} \\
\addlinespace[2pt]
\multirow{2}{*}{\texttt{gpt-4o}}
 & \textbf{1.414} \scriptsize$\pm$0.07 & 1.342 \scriptsize$\pm$0.09 & 1.275 \scriptsize$\pm$0.16 & 1.139 \scriptsize$\pm$0.11 & 1.211 \scriptsize$\pm$0.13 & 1.188 \scriptsize$\pm$0.09 & 1.032 \scriptsize$\pm$0.17 & \underline{1.281} \scriptsize$\pm$0.17 & 1.342 \scriptsize$\pm$0.11 & 1.247 \\
 & (86.7\%) & (86.4\%) & (84.6\%) & (78.9\%) & (84.6\%) & (84.4\%) & (78.6\%) & (83.9\%) & (87.4\%) & (83.9\%) \\
\cmidrule(lr){1-11} 
\multirow{2}{*}{\texttt{gpt-4o-mini}}
 & 1.164 \scriptsize$\pm$0.11 & 1.228 \scriptsize$\pm$0.12 & \underline{1.402} \scriptsize$\pm$0.11 & \underline{1.230} \scriptsize$\pm$0.14 & \textbf{1.381} \scriptsize$\pm$0.11 & 1.105 \scriptsize$\pm$0.16 & \textbf{1.376} \scriptsize$\pm$0.19 & 0.973 \scriptsize$\pm$0.16 & 1.305 \scriptsize$\pm$0.17 & 1.240 \\
 & (77.1\%) & (74.4\%) & (89.7\%) & (79.6\%) & (89.9\%) & (78.9\%) & (88.5\%) & (56.0\%) & (76.7\%) & (79.0\%) \\
\cmidrule(lr){1-11} 
\multirow{2}{*}{\texttt{gpt-3.5-turbo}}
 & \underline{1.322} \scriptsize$\pm$0.11 & \underline{1.366} \scriptsize$\pm$0.12 & 1.344 \scriptsize$\pm$0.14 & 1.159 \scriptsize$\pm$0.07 & 1.159 \scriptsize$\pm$0.07 & \underline{1.254} \scriptsize$\pm$0.08 & 1.177 \scriptsize$\pm$0.12 & 1.140 \scriptsize$\pm$0.12 & \underline{1.395} \scriptsize$\pm$0.18 & \underline{1.257} \\
 & (94.4\%) & (90.0\%) & (90.5\%) & (81.3\%) & (95.8\%) & (90.7\%) & (84.6\%) & (78.2\%) & (91.0\%) & (88.5\%) \\
\midrule
\multicolumn{11}{l}{\textit{\textbf{Gemini-series}}} \\
\addlinespace[2pt]
\multirow{2}{*}{\texttt{gemini-1.5-pro}}
 & 1.202 \scriptsize$\pm$0.15 & \textbf{1.394} \scriptsize$\pm$0.19 & \textbf{1.502} \scriptsize$\pm$0.09 & \textbf{1.350} \scriptsize$\pm$0.19 & \underline{1.324} \scriptsize$\pm$0.13 & \textbf{1.522} \scriptsize$\pm$0.09 & \underline{1.308} \scriptsize$\pm$0.13 & \textbf{1.300} \scriptsize$\pm$0.40 & \textbf{1.525} \scriptsize$\pm$0.18 & \textbf{1.381} \\
 & (72.0\%) & (75.8\%) & (93.6\%) & (74.6\%) & (66.7\%) & (89.4\%) & (94.2\%) & (68.0\%) & (88.8\%) & (80.3\%) \\
\cmidrule(lr){1-11} 
\multirow{2}{*}{\texttt{gemini-1.5-flash}}
 & 0.875 \scriptsize$\pm$0.23 & 1.005 \scriptsize$\pm$0.21 & 1.304 \scriptsize$\pm$0.15 & 0.951 \scriptsize$\pm$0.17 & 1.110 \scriptsize$\pm$0.20 & 0.988 \scriptsize$\pm$0.20 & 1.056 \scriptsize$\pm$0.10 & 0.826 \scriptsize$\pm$0.18 & 1.284 \scriptsize$\pm$0.17 & 1.044 \\
 & (61.8\%) & (66.3\%) & (87.6\%) & (61.8\%) & (65.0\%) & (71.8\%) & (75.4\%) & (54.4\%) & (82.6\%) & (69.6\%) \\
\bottomrule
\end{tabular}
}
\vspace{-7pt}
\caption{Buyer \textsc{Merit} performance with standard deviation using the ReAct \cite{yao2023react} framework across different market conditions. Each cell presents the \textsc{Merit} score (top) and deal rate (bottom, in parentheses). \textbf{Bold} indicates the best performance, and \underline{underline} indicates the second best.}
\vspace{-15pt}
\label{tab:buyer_performance}
\end{table*}

\renewcommand{\arraystretch}{1.0}

\paragraph{Negotiation power}

Bargaining outcomes also hinge on the buyer’s ability to \textit{shift} the final price in their favor—an aspect tied to the broader economic concept of \textit{bargaining power} \cite{chatterjee1983bargaining, muthoo1999bargaining}. \textsc{Merit} incorporates this through:
\vspace{-2pt}
\begin{equation}
\text{Negotiation Power (NP)} = \frac{P_{\text{initial}}-P_{\text{deal}}}{P_{\text{initial}}-P_{\text{cost}}}
\label{eqs:np}
\end{equation}
where  $P_{\text{initial}}$  is the seller’s initial asking price. A higher NP indicates that the buyer successfully negotiated the deal price downward from the initial offer. In classic bargaining theory (e.g., Nash bargaining \cite{binmore1986nash}), each party’s utility share depends on their relative power. Here, NP complements the \textit{cardinal utility} of consumer surplus by capturing how effectively the buyer reclaims surplus that might otherwise go to the seller. Values near 1 imply maximal movement from the seller’s initial price, reflecting strong buyer influence, whereas values near 0 imply that the buyer gained little ground in negotiations.

\paragraph{Acquisition ratio of desired goods}
Finally, a consumer’s satisfaction also depends on obtaining the \textit{right} product, reflecting the ordinal aspect of utility—that is, how closely the outcome aligns with the consumer’s preference ranking. We quantify this via an \textit{Acquisition Ratio}, which measures how semantically similar the buyer’s acquired items are to their desired items. By embedding items in a high-dimensional space using \texttt{text-embedding-ada-002}, we compute:

\begin{equation}
\text{Acquisition Ratio (AR)} = \frac{\mathbf{v_{\text{acquired}}} \cdot \mathbf{v_{\text{desired}}}}{\|\mathbf{v_{\text{acquired}}}\| \|\mathbf{v_{\text{desired}}}\|}
\end{equation}
where $\mathbf{v_{\text{acquired}}}$ and $\mathbf{v_{\text{desired}}}$ represent embedding vectors of the items actually purchased and initially sought, respectively. A higher AR indicates that the buyer’s outcome closely matches their \textit{top preferences}, embodying the idea that utility is also determined by which specific goods are consumed, not merely how profitable the purchase is. 

\subsection{Human preference alignment}
\label{subsec:human_preference_alignment}

\begin{figure}[t]
\centering
\small
\centerline{\includegraphics[width=1.0\columnwidth]{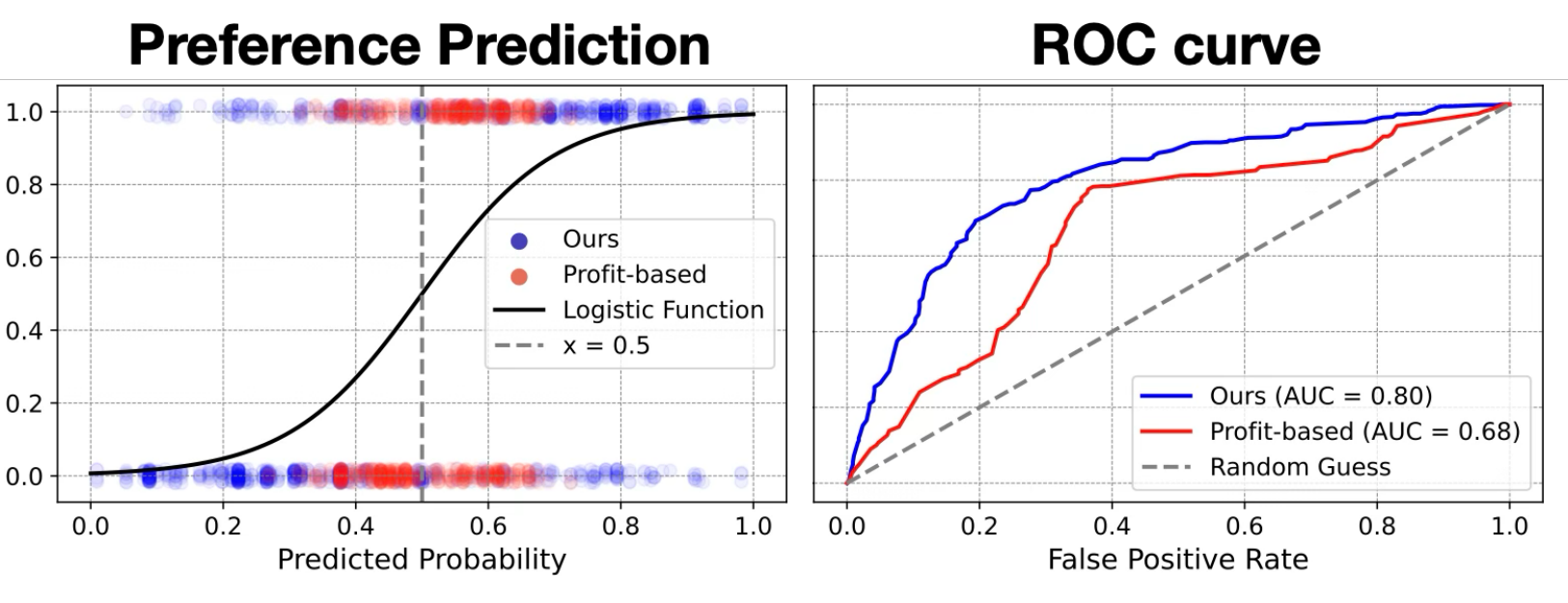}}
\vspace{-10pt}
\caption{Demonstration that a human-aligned metric outperforms a purely profit-based one.}
\label{fig:human_preference}
\vspace{-15pt}
\end{figure}

\noindent
To ensure \textsc{Merit} reflects human judgments, we ran an Amazon Mechanical Turk (MTurk) study. Participants were shown pairs of negotiation dialogues (20 comparisons per product; 180 participants total) and asked to choose the better one. The survey included both same-item price comparisons and trade-offs between similar items; full details are in Appendix~\ref{app:sec:human_preference}. 

The coefficients $\alpha, \beta, \gamma$ in \autoref{eq:profit_equation} were optimized globally using merged data from all product categories based on Bradley-Terry model \cite{Bradley1952RankAnalysis}. Since only relative magnitudes matter in the Bradley-Terry model, we scale them for interpretability so that their sum is 3, as follows:
\[
\scriptstyle \alpha' = \frac{\alpha}{\alpha+\beta+\gamma} \times 3,\quad 
\beta' = \frac{\beta}{\alpha+\beta+\gamma} \times 3,\quad 
\gamma' = \frac{\gamma}{\alpha+\beta+\gamma} \times 3.
\]
After optimization, the final scaled coefficients are: $\alpha' = 1.0139$, $\beta' = 0.8812$, $\gamma' = 1.1049$.

As shown in \autoref{fig:human_preference}, this process confirmed that \textsc{Merit} aligns more closely with human preferences than traditional profit-based ones (e.g., achieving an ROC AUC of 0.80 vs. 0.68 for the latter). This human-aligned metric is subsequently used for model evaluation and as feedback in our system. Details on the optimization of the metric model are provided in Appendix \ref{app:sec:human_preference}.
\section{Behaviors of LLM Negotiators}
\label{sec:behaviors}

\begin{table}[t]
\centering
\resizebox{\columnwidth}{!}{%
\begin{tabular}{lcccccc}
\toprule
  & \multicolumn{2}{c}{\textbf{Market 1$^{*}$}}  & \multicolumn{2}{c}{\textbf{Market 2$^{**}$}} & \multicolumn{2}{c}{\textbf{Market 3$^{***}$}}  \\ 
  \cmidrule(lr){2-3} \cmidrule(lr){4-5} \cmidrule(lr){6-7} 
\textbf{Buyers}    & \textbf{AVG price} \textcolor{red}{$\downarrow$} & \textbf{\textsc{Merit}} \textcolor{blue}{$\uparrow$} & \textbf{AVG price} \textcolor{red}{$\downarrow$}& \textbf{\textsc{Merit}} \textcolor{blue}{$\uparrow$} & \textbf{AVG price} \textcolor{red}{$\downarrow$}& \textbf{\textsc{Merit}} \textcolor{blue}{$\uparrow$}\\ 
\midrule
\texttt{gemini-1.5-pro}       & 444.6 & 2.18 & 432.1  & 2.38  & 422.6  & 2.39\\ 
\texttt{gemini-1.0-pro}       & 470.2 & 1.63 & 455.4  & 2.05 & 434.7  & 2.11\\ 
\texttt{gemini-1.5-flash}     & 481.4 & 1.62 & 476.0 & 1.68 & 485.6 & 1.61\\ 
\bottomrule
\multicolumn{7}{l}{*: initial price 550$\,\,\,\,$ **: initial price 520$\,\,\,\,$ ***: Buyer propose the price first.}
\end{tabular}
}
\vspace{-5pt}
\caption{Anchoring effect and first-mover advantage.}
\label{tab:anchoring}
\vspace{-15pt}
\end{table}

In this section, we evaluate the negotiation capabilities of contemporary LLMs, specifically the \texttt{GPT-series} and \texttt{Gemini-series}, within our \textsc{AgoraBench}. The experiments are designed to uncover emergent behaviors and provide insights into LLM performance across varied strategic contexts.

\paragraph{Anchoring effect and first-mover advantage (\autoref{tab:anchoring})}
We investigated the \textit{anchoring effect} \citep{orr2005anchoring}, a cognitive bias where initial information heavily influences subsequent decisions. Experiments in the vanilla market, using two distinct initial camera prices (\$550 and \$520, denoted Market 1$^{*}$ and Market 2$^{**}$ respectively) set by the seller, demonstrated this bias: higher initial anchor prices generally led to higher final deal prices. Furthermore, when the buyer initiated the first offer (Market 3$^{***}$), the average deal price was even lower compared to scenarios where the seller proposed first, highlighting a potential first-mover advantage for the buyer in price setting.

\paragraph{Irrational concessions: Unstable anchoring (\autoref{tab:buyer_seller_camera_negotiation_ex})}
\label{sec:irrational_concessions}

We observed unnatural negotiation behavior in LLMs, especially smaller models such as \texttt{gpt-4o-mini}. In \autoref{tab:buyer_seller_camera_negotiation_ex}, the buyer shows unstable anchoring: after proposing a price, it reverses direction with strictly lower counteroffers, effectively backtracking on its stated stance. Concretely, the buyer drops from \$480 to \$475 and \$450, then later returns to \$480 before agreement. Such reversed concessions are atypical in human bargaining, where offers are generally monotone toward settlement, and they weaken the buyer’s anchor by signaling inconsistent \textit{willingness to pay} reflecting anchor instability rather than strategic bargaining.

\paragraph{Impact of market conditions (\autoref{tab:buyer_performance})}
\label{sec:market_condition_effects}

\noindent
\begin{itemize}[leftmargin=*, itemsep=2pt, topsep=2pt]
    \item \textbf{Deceptive.} Allowing buyers to use deception generally \emph{improves buyer outcomes}: \textsc{Merit} increases vs.\ vanilla (avg.\ +0.10) and deal rates rise substantially (+10.62\%p). Some GPT models trade slightly lower deal rates for higher scores (e.g., \texttt{gpt-4o}), whereas Gemini often improves both, suggesting a clearer buyer advantage.
    \item \textbf{Monopoly.} Monopoly consistently harms buyers across models: deal rates drop by 3.34\% (single-item) and 8.80\% (multi item), while \textsc{Merit} declines by about 0.10 and 0.13, respectively. This reflects increased seller leverage, higher prices, and weaker convergence.
    \item \textbf{Installment.} Installment options have mixed effects: in single-item settings they often increase deal rates but at the cost of worse buyer prices (lower \textsc{Merit}); in multi-item settings, added bargaining complexity tends to reduce deal rates overall.
    \item \textbf{Negative perception.} A negative prior about the seller consistently reduces deal rates, with a sharper effect in single-item negotiations (few outside options). In multi-item settings, substitutes provide limited mitigation but do not eliminate the decline.
    \item \textbf{Multi product.}
Across conditions, multi-item settings generally exhibit higher deal rates than single-item ones, consistent with the presence of substitutes that facilitate agreement. However, this benefit is not uniform: under \textit{installment}, GPT models show a reversal where deal rates drop despite substitutes. Such sign reversals serve as a useful diagnostic of market compatibility, indicating failures to leverage multi product option when bargaining becomes multi-dimensional.
\end{itemize}

\noindent
By comparing each market condition against the benchmark’s baseline (\textit{deceptive} or vanilla) in terms of both deal rate and buyer \textsc{Merit}, we can assess whether LLM's \emph{market compatibility} they align with patterns typical in human bargaining or deviates from them. In particular, directional shifts across conditions (e.g., reduced agreement and utility under \textit{monopoly}, or trust-driven drops under \textit{negative perception}) reveal whether a model responds to changes in leverage, multiple options, and priors in an economically coherent manner. For detailed explanation of LLMs behaviors, see in Appendix \ref{app:sec:behaviors_of_llms}.
\section{\textsc{Merit}-Guided Prompting \& Training}
\label{sec:ICL_UF}

To address the common misalignment of LLM-based negotiations with human preferences, we propose \textsc{Merit}-guided In-Context Learning (ICL-MF) and training to enhance negotiation strategies.

\begin{table}[t]
  \centering
  \small
  % \vspace{-7pt}
  \setlength{\tabcolsep}{4pt}
  \renewcommand{\arraystretch}{1.00}
  
  \newcommand{\twolines}[2]{%
    {\renewcommand{\arraystretch}{0.85}%
    \begin{tabular}[t]{@{}c@{}}#1\\[-0.45ex]{\scriptsize #2}\end{tabular}}%
  }
  \newcommand{\modelcell}[1]{%
    {\renewcommand{\arraystretch}{0.85}%
    \begin{tabular}[t]{@{}l@{}}#1\\[-0.45ex]\phantom{\scriptsize DR: 100.0}\end{tabular}}%
  }

  \resizebox{\linewidth}{!}{%
    \begin{tabular}{l| c c c c c c}
      \toprule
      & \multicolumn{3}{c}{Single product}
      & \multicolumn{3}{c}{Multi product} \\
      \cmidrule(lr){2-4} \cmidrule(lr){5-7}
      Model (vs 1.5-pro)
        & ReAct & OG-Narrator & Ours
        & ReAct & OG-Narrator & Ours \\
      \midrule

      \modelcell{GPT-4o}
        & \twolines{1.127}{84.2\%}
        & \twolines{1.094}{51.6\%}
        & \cellcolor{pink!30}\textbf{\twolines{1.662}{100\%}}
        & \twolines{1.045}{76.1\%}
        & \twolines{1.478}{82.8\%}
        & \cellcolor{pink!30}\textbf{\twolines{1.563}{99.7\%}} \\

      \modelcell{GPT-4o-mini}
        & \twolines{1.143}{73.0\%}
        & \twolines{1.165}{45.1\%}
        & \cellcolor{pink!30}\textbf{\twolines{1.450}{97.1\%}}
        & \twolines{1.493}{96.4\%}
        & \twolines{1.281}{66.1\%}
        & \cellcolor{pink!30}\textbf{\twolines{1.527}{98.6\%}} \\

      \modelcell{GPT-3.5-turbo}
        & \twolines{1.248}{86.9\%}
        & \twolines{1.157}{45.5\%}
        & \cellcolor{pink!30}\textbf{\twolines{1.510}{99.1\%}}
        & \twolines{1.119}{82.8\%}
        & \twolines{1.346}{71.9\%}
        & \cellcolor{pink!30}\textbf{\twolines{1.456}{100\%}} \\

      \modelcell{Gemini-1.5-pro}
        & \twolines{1.354}{76.0\%}
        & \twolines{1.197}{43.8\%}
        & \cellcolor{pink!30}\textbf{\twolines{1.841}{99.8\%}}
        & \twolines{1.638}{97.2\%}
        & \twolines{1.544}{81.9\%}
        & \cellcolor{pink!30}\textbf{\twolines{1.768}{100\%}} \\

      \modelcell{Gemini-1.5-flash}
        & \twolines{0.929}{63.2\%}
        & \twolines{0.689}{26.9\%}
        & \cellcolor{pink!30}\textbf{\twolines{1.418}{99.8\%}}
        & \twolines{1.506}{98.9\%}
        & \twolines{1.422}{80.3\%}
        & \cellcolor{pink!30}\textbf{\twolines{1.565}{99.7\%}} \\
        \midrule
      \modelcell{Average}
      & \twolines{1.160}{76.7\%}
      & \twolines{1.060}{42.6\%}
      & \cellcolor{pink!30}\textbf{\twolines{1.576}{99.2\%}}
      & \twolines{1.360}{90.3\%}
      & \twolines{1.414}{76.6\%}
      & \cellcolor{pink!30}\textbf{\twolines{1.576}{99.6\%}} \\

      \midrule
      \multicolumn{7}{l}{Model (vs DeepSeek-v3.2)} \\
      \midrule
      \modelcell{GPT-5.1}
  & \twolines{1.849}{99.6\%}
  & \twolines{1.423}{50.4\%}
  & \cellcolor{pink!30}\textbf{\twolines{2.027}{99.1\%}}
  & \twolines{1.950}{100\%}
  & \twolines{1.536}{85.8\%}
  & \cellcolor{pink!30}\textbf{\twolines{2.095}{99.7\%}} \\

\modelcell{DeepSeek-v3.2}
  & \twolines{1.675}{100\%}
  & \twolines{0.333}{12.2\%}
  & \cellcolor{pink!30}\textbf{\twolines{1.875}{98.9\%}}
  & \twolines{1.628}{100\%}
  & \twolines{0.957}{52.5\%}
  & \cellcolor{pink!30}\textbf{\twolines{1.872}{100\%}} \\

\modelcell{Sonnet 4.5}
  & \twolines{1.457}{99.8\%}
  & \cellcolor{blue!15}\textbf{\twolines{1.856}{67.1\%}}
  & \twolines{1.778}{99.8\%}
  & \twolines{1.560}{100\%}
  & \twolines{1.368}{53.1\%}
  & \cellcolor{pink!30}\textbf{\twolines{1.890}{100\%}} \\

\modelcell{Gemini-3-Flash}
  & \twolines{1.870}{100\%}
  & \cellcolor{blue!15}\textbf{\twolines{2.116}{76.7\%}}
  & \twolines{1.995}{99.8\%}
  & \twolines{1.856}{100\%}
  & \twolines{1.346}{56.4\%}
  & \cellcolor{pink!30}\textbf{\twolines{2.035}{99.7\%}} \\

\modelcell{gpt-oss-120b}
  & \twolines{1.318}{89.6\%}
  & \twolines{0.098}{4.9\%}
  & \cellcolor{pink!30}\textbf{\twolines{1.391}{90.6\%}}
  & \twolines{1.268}{86.4\%}
  & \twolines{0.934}{60.3\%}
  & \cellcolor{pink!30}\textbf{\twolines{1.431}{90.3\%}} \\

\modelcell{gpt-oss-20b}
  & \twolines{1.325}{89.3\%}
  & \twolines{0.154}{5.6\%}
  & \cellcolor{pink!30}\textbf{\twolines{1.413}{91.3\%}}
  & \twolines{0.986}{65.3\%}
  & \twolines{0.545}{33.8\%}
  & \cellcolor{pink!30}\textbf{\twolines{1.410}{86.6\%}} \\

\midrule
\modelcell{Average}
  & \twolines{1.582}{96.4\%}
  & \twolines{0.997}{36.2\%}
  & \cellcolor{pink!30}\textbf{\twolines{1.747}{96.6\%}}
  & \twolines{1.541}{92.0\%}
  & \twolines{1.114}{57.0\%}
  & \cellcolor{pink!30}\textbf{\twolines{1.789}{96.1\%}} \\

  \midrule  
  \multicolumn{7}{l}{Model (vs DeepSeek-v3.2)} \\
  \midrule
  \modelcell{GPT-5.1 reasoning mode}
  & \twolines{1.999}{100\%}
  & \twolines{1.315}{46.9\%}
  & \cellcolor{pink!30}\textbf{\twolines{2.127}{100\%}}
  & \twolines{2.055}{100\%}
  & \twolines{1.452}{78.9\%}
  & \cellcolor{pink!30}\textbf{\twolines{2.189}{99.7\%}} \\

\modelcell{DeepSeek-v3.2 reasoning mode}
  & \twolines{1.648}{99.8\%}
  & \twolines{0.275}{10.4\%}
  & \cellcolor{pink!30}\textbf{\twolines{1.852}{98.9\%}}
  & \twolines{1.632}{99.2\%}
  & \twolines{0.900}{49.2\%}
  & \cellcolor{pink!30}\textbf{\twolines{1.885}{99.4\%}} \\

\midrule
\modelcell{Average}
  & \twolines{1.824}{99.9\%}
  & \twolines{0.795}{28.7\%}
  & \cellcolor{pink!30}\textbf{\twolines{1.990}{99.5\%}}
  & \twolines{1.844}{99.6\%}
  & \twolines{1.176}{64.1\%}
  & \cellcolor{pink!30}\textbf{\twolines{2.037}{99.6\%}} \\

      \bottomrule
      
    \end{tabular}%
  }
  \vspace{-5pt}
  \caption{We compare ICL-MF with ReAct and OG-Narrator using \textsc{Merit} and deal rate. We run 10 dialogues per product under each market condition, and report averages separately for the single-item and multi-item settings. Cells are highlighted in \colorbox{pink!30}{pink} when ICL-MF achieves the best performance, and in \colorbox{blue!15}{light purple} when OG-Narrator achieves the best performance.}
  \label{tab:total_results}
  \vspace{-10pt}
\end{table}

\begin{figure}[t]
\centering
\small
\centerline{\includegraphics[width=0.9\columnwidth]{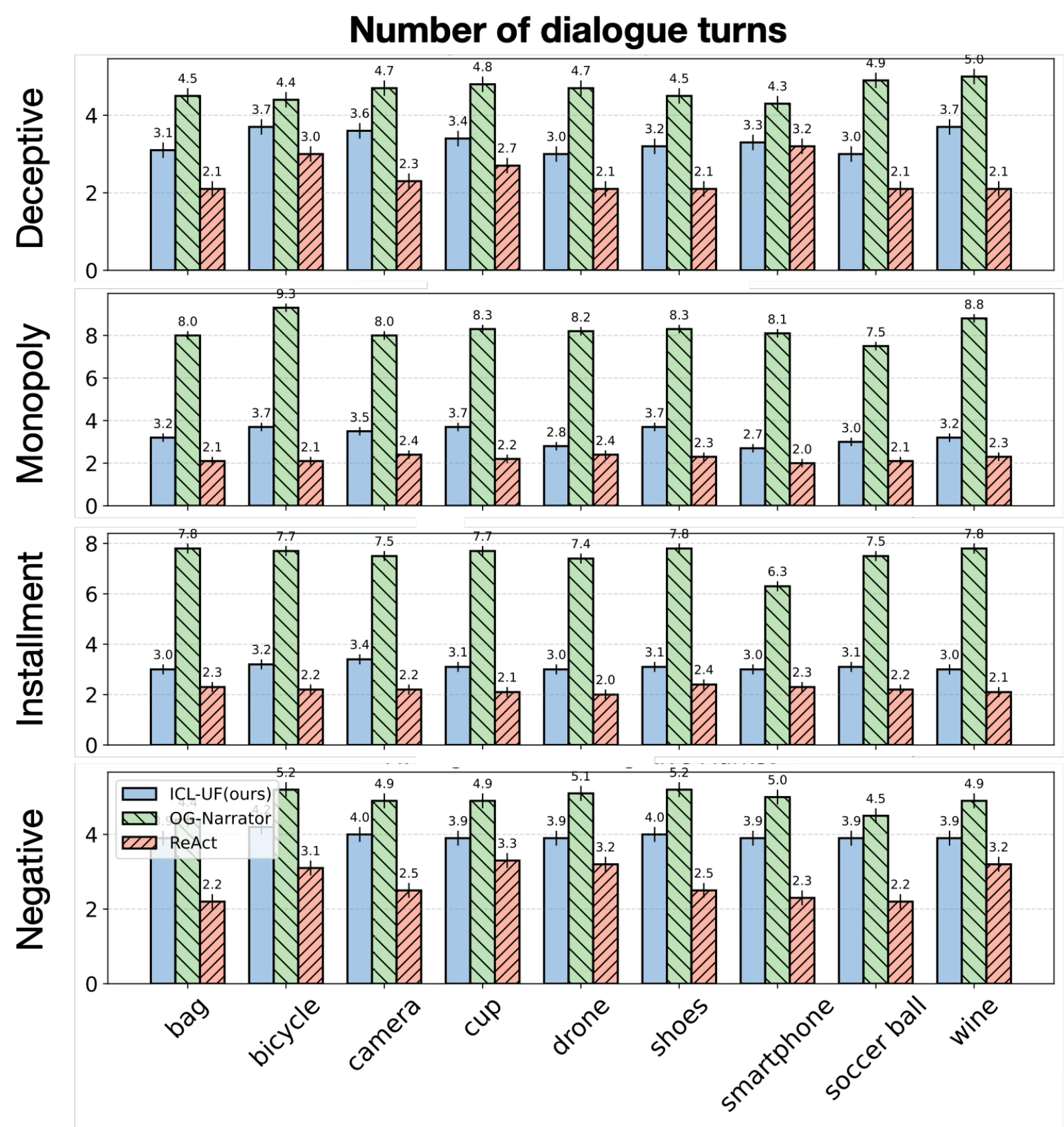}}
\vspace{-10pt}
\caption{Number of turns for the negotiation. We feed dialogues from \texttt{gpt-4o} buyer and \texttt{gemini-1.5-pro} seller to \texttt{gemma-3-27b} judge.}
\label{fig:number_of_turns}
\vspace{-10pt}
\end{figure}

\subsection{Human preference evaluation (\textsc{Merit})}
\paragraph{\textsc{Merit}-guided Prompting over Others.}
We compared ICL-MF with two baselines; ReAct \cite{yao2023react}, which prompts LLMs to reason before acting, and OG-Narrator \cite{xia2024measuring}, which generates offers by linearly interpolating the buyer’s price from 50\% to 100\% of the budget across turns. We evaluate the performance with \textsc{Merit} for five buyer (closed-sourced) LLMs (GPT/Gemini series) negotiating against a \texttt{gemini-1.5-pro} seller. Further, to assess scalability across model families, we evaluate six buyer LLMs—GPT-5.1 \cite{singh2025openai}, DeepSeek-v3.2 \cite{liu2025deepseek}, Sonnet 4.5 \cite{anthropic2025claude45systemcard}, Gemini-3-Flash \cite{deepmind_gemini_models_page} (closed-source) and gpt-oss-120b, gpt-oss-20b (open-source; the gpt-oss series \cite{agarwal2025gpt})—against a \texttt{DeepSeek-v3.2} seller. Also we conduct experiments based on the reasoning model (switching on the reasoning mode of  GPT-5.1 and DeepSeek-v3.2) to observe whether performance gain exist in the reasoning mode. As shown in \autoref{tab:total_results}, ICL-MF consistently outperforms both baselines by a large margin in terms of \textsc{Merit} and deal rate in both single- and multi-product settings, and these gains hold consistently across closed-sourced models, open-sourced models, and reasoning variants. 

\begin{table*}[ht]
\centering
\footnotesize
\renewcommand{\arraystretch}{0.90} 
\begin{tabularx}{\linewidth}{>{\raggedright\arraybackslash}p{1.8cm} | X}
\toprule
\textbf{Method} & \textbf{Exemplary Thought Process} \\ \midrule
\textbf{ReAct} & \textit{Thought1: "\$475 is still a bit high. I'll push a little further and feign disinterest to see if they'll budge."} \\ \cmidrule{2-2}
 & \textit{Thought2: "\$550 is above my budget. I'll start low and express hesitation... I'll hint I might walk away to add pressure."} \\ \midrule
\textbf{ICL-MF} & \textit{Thought1: "\underline{They moved a decent amount, suggesting their cost is probably closer to \$400.} If their cost is \$400, a deal at \$450 would give me $CS=0.5$ and $NP=0.67$."}\\ \cmidrule{2-2}
  & \textit{Thought2: \underline{"I'm guessing they might be willing to go a bit lower—maybe their cost is closer to \$350.} At \$450, $CS$ would be 0.33 and $NP$ would be 0.5. I need to make them think I might walk away."} \\ \bottomrule
\end{tabularx}
\vspace{-7pt}
\caption{Comparison of internal reasoning: ReAct vs. ICL-MF with \texttt{gemini-1.5-pro}. While ReAct uses qualitative tactics, ICL-MF quantifies utility through explicit cost estimation and economic metrics ($CS, NP$).}
\vspace{-7pt}
\label{tab:reasoning_comparison}
\end{table*}

\begin{table}[t]
  \centering
  \small
  \setlength{\tabcolsep}{4pt}
  \renewcommand{\arraystretch}{0.85}
  \newcommand{\twolines}[2]{%
    {\renewcommand{\arraystretch}{0.65}%
    \begin{tabular}[t]{@{}c@{}}#1\\[-0.75ex]{\scriptsize #2}\end{tabular}}%
  }
  \newcommand{\modelcell}[1]{%
    {\renewcommand{\arraystretch}{0.65}%
    \begin{tabular}[t]{@{}l@{}}#1\\[-0.75ex]\phantom{\scriptsize DR: 100.0}\end{tabular}}%
  }

  \resizebox{\linewidth}{!}{%
    \begin{tabular}{l| c c c c}
      \toprule
      & \multicolumn{2}{c}{Single product}
      & \multicolumn{2}{c}{Multi product} \\
      \cmidrule(lr){2-3} \cmidrule(lr){4-5}
      gpt-oss-20b (vs DeepSeek-v3.2)
        & \textsc{Merit} & Deal rate
        & \textsc{Merit} & Deal rate \\
      \midrule

      \modelcell{ReAct}
        & 1.325 & 89.3\% & 0.986 & 65.3\% \\[-0.35ex]

      \modelcell{ICL-MF (ours)}
        & \textbf{1.413} & \textbf{91.3\%} & \underline{1.410} & \underline{86.6\%} \\[-0.35ex]

      \modelcell{SFT (ours)}
        & \underline{1.332} & \underline{89.6\%} & \textbf{1.459} & \textbf{93.1\%} \\[-0.35ex]

      \bottomrule
    \end{tabular}%
  }
  \vspace{-5pt}
  \caption{We compare ReAct with our data-utilized methods; ICL and SFT utilizing gpt-oss-20b. \textbf{Bold} indicates the best performance, and \underline{underline} indicates the second best.}
  \label{tab:sft_results}
  \vspace{-7pt}
\end{table}

\noindent
\textbf{Analysis of results.} OG-Narrator often underperforms ReAct due to its rigid price schedule, which fails to adapt to opponent temperament. Against patient sellers like \texttt{gemini-1.5-pro}, it excessively prolongs dialogue (\autoref{fig:number_of_turns}), leading to bargaining impasses \cite{babcock1997explaining, hwang2018theory, backus2020communication}, without reaching a deal. Conversely, with impatient sellers like \texttt{deepseek-v3.2}, negotiations terminate abruptly as opponents exit prematurely. While ReAct settles too quickly due to premature assumptions \cite{laban2025llms}, ICL-MF balances efficiency and effectiveness through appropriate negotiation lengths. Crucially, ICL-MF demonstrates a clear positive correlation between model capability and performance, better leveraging the inherent strengths of stronger LLMs.

We additionally report raw economic outcomes beyond \textsc{Merit}: \textit{Deal Rate}, \textit{Average Deal Price (deals only)}, and \textit{Average Buyer Profit}. \textit{Average Deal Price} is computed over successful negotiations only and should be interpreted jointly with \textit{Deal Rate}. As shown in \autoref{tab:multi_eval_single} (Appendix~\ref{app:sec:single_eval_pathway}), these complementary metrics are consistent with our main results, alleviating the \textit{single evaluation pathway} concern.

\noindent
\paragraph{\textsc{Merit} elicits opponent-aware reasoning.} Adapting \textsc{Merit} as guidance for negotiation encourages LLMs to transition from simple tactic-centric thoughts to deep opponent-aware reasoning (OAR). As illustrated in \autoref{tab:reasoning_comparison}, while the ReAct baseline relies on vague behavioral maneuvers such as ``feigning disinterest" or ``adding pressure", ICL-MF enables the agent to explicitly hypothesize the seller's hidden belief, specifically their underlying cost. For instance, the ICL-MF agent deduces that the opponent's cost is likely ``closer to \$400'' based on their previous moves, subsequently calculating its own Consumer Surplus (CS) and Negotiation Power (NP) to evaluate the feasibility of a deal. 

\paragraph{Human preference dataset training}
We enhance the negotiation power of LLMs by fine-tuning \texttt{gpt-oss-20b} using LoRA on human-preferred dialogues (\autoref{fig:main} (b)). To maintain practical realism, we exclude unobservable seller thoughts from the training data. We define a filtered preferred dialogue as a turn sequence $\mathcal{D} = \{(b^{r}_t, b^{u}_t, b^{a}_t, s^{u}_t, s^{a}_t)\}_{t=0}^{T-1}$, where $b^{r}_t$, $b^{u}_t$, and $b^{a}_t$ denote the buyer's reasoning, utterance, and action at turn $t$, respectively. The model is trained to predict the buyer's next turn $y_t$ given the interaction history $x_t$:
\begin{equation}
\begin{aligned}
b_t &= (b^{r}_t, b^{u}_t, b^{a}_t), \quad s_t = (s^{u}_t, s^{a}_t) \\
x_t &= \big[\, b_0, s_0, b_1, s_1, \dots, s_{t-1} \,\big], \\
y_t &= b_t, \quad t = 0, 1, \dots, T-1
\end{aligned}
\end{equation}
The buyer policy $\pi_\theta$ is optimized using the standard maximum-likelihood objective:
\begin{equation}
\mathcal{L}_{\mathrm{SFT}}(\theta) = - \sum_{\mathcal{D}} \sum_{t=0}^{T-1} \log \pi_\theta(y_t \mid x_t)
\end{equation}

As shown in \autoref{tab:sft_results}, the SFT \texttt{gpt-oss-20b} consistently outperforms the ReAct baseline when simulated against \texttt{deepseek-v3.2}. Notably, while ICL-MF excels in single-product settings due to efficient situational matching, SFT proves superior in multi-product negotiations that require stable management of diverse products and complex trade-offs. These results demonstrate that leveraging human negotiation strategies leads to more robust outcomes than pure reasoning-based approaches.

Importantly, the SFT model is trained only in the Deceptive regime with GPT-/Gemini-family opponents but evaluated across all nine regimes and against an unseen \texttt{DeepSeek-v3.2} seller, constituting a dual out-of-distribution shift.

\begin{figure}[t]
\centering
\small
\centerline{\includegraphics[width=1.0\columnwidth]{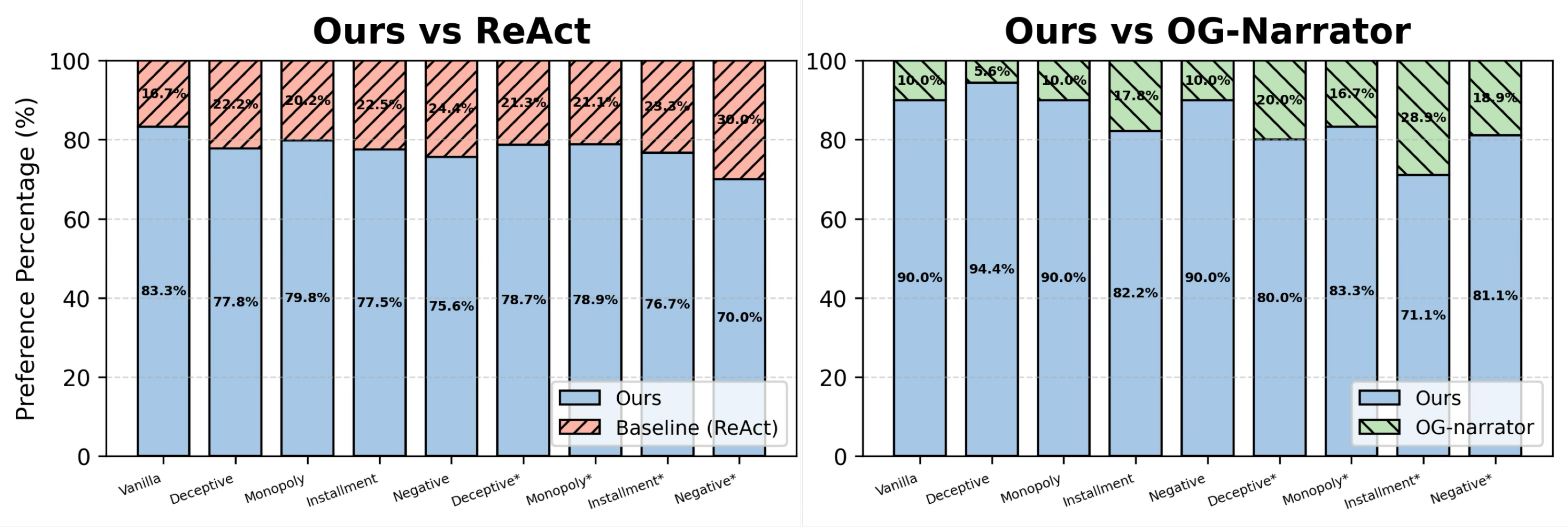}}
\vspace{-10pt}
\caption{Preference comparison between ICL-MF vs ReAct; OG-Narrator algorithm by LLM-judge}
\label{fig:ours vs standard preference}
\vspace{-15pt}
\end{figure}

\subsection{LLM preference evaluation}
We also assessed generated negotiation dialogues using an LLM as an impartial judge. This aligns with the emerging paradigm of leveraging LLMs for fine-grained evaluation \cite{zheng2023judging, chiang2023vicuna}, offering a scalable means to capture interaction aspects that automated scores might overlook. For this task, gemma-3-27b-it \cite{gemma_2025} evaluated dialogues from ICL-MF (\texttt{gpt-4o} buyer vs. \texttt{gemini-1.5-pro} seller) against those from two baselines: ReAct and OG-Narrator. The LLM-judge was presented with paired dialogues in random order and prompted to select the negotiation more favorable to the buyer, considering outcome, budget adherence, and product features (see Appendix \ref{app:sec:system prompt} for the detailed prompt).

The results in \autoref{fig:ours vs standard preference} show a clear preference for ICL-MF across all market conditions. Against OG-Narrator, ICL-MF often exceeded 84.7\% preference on average, peaking at 94.4\% in the deceptive market and remaining strong (e.g., 71.3\%) even under installment scenarios. Against ReAct, ICL-MF likewise outperformed with preference rates typically between 70\% and over 83.3\%. Overall, this independent LLM-judge evaluation corroborates our main findings: ICL-MF not only optimizes the targeted utility components but also produces buyer-perceived strategies and outcomes that are more favorable and strategically sound.
\section{Conclusion}
We introduce a dataset and benchmark that mirror economically grounded market complexity, a human-aligned multi-facet metric, and an in-context learning and training method that uses this metric as feedback to enhance negotiation skills. By capturing both profit and acquisition priorities, our framework fills gaps left by profit-only metrics and, we hope, helps the community build more effective, human-aligned LLM bargaining agents.

\section*{Limitations}
While our work advances LLM bargaining, several limitations suggest avenues for future exploration. First, our study is primarily buyer-centric, overlooking seller-specific objectives such as inventory risk or tiered profit targets; future work should incorporate a dedicated seller perspective. Second, while \textsc{AgoraBench} covers diverse scenarios, it does not yet capture global market subtleties like regulatory contexts or cultural norms. Third, we do not cover tool-augmented agentic environments (e.g., LLM-agent tool use for local retrieval or search), which could materially affect bargaining behavior by enabling grounded, real-time access to external information (e.g., comparable prices or product specifications). While building such environments would be valuable, it is beyond the scope of this study; we leave tool-integrated settings (including local search) to future work.

\bibliography{custom}

@article{lewis2017deal,
  title={Deal or no deal? end-to-end learning for negotiation dialogues},
  author={Lewis, Mike and Yarats, Denis and Dauphin, Yann N and Parikh, Devi and Batra, Dhruv},
  journal={arXiv preprint arXiv:1706.05125},
  year={2017}
}

@article{he2018decoupling,
  title={Decoupling strategy and generation in negotiation dialogues},
  author={He, He and Chen, Derek and Balakrishnan, Anusha and Liang, Percy},
  journal={arXiv preprint arXiv:1808.09637},
  year={2018}
}

@article{mankiw2007principles,
  title={Principles of macroeconomics},
  author={Mankiw, N Gregory and Kneebone, Ronald D and McKenzie, Kenneth James and Rowe, Nicholas},
  year={2007},
  publisher={Thomson South-Western Mason, OH}
}

@article{fu2023improving,
  title={Improving language model negotiation with self-play and in-context learning from ai feedback},
  author={Fu, Yao and Peng, Hao and Khot, Tushar and Lapata, Mirella},
  journal={arXiv preprint arXiv:2305.10142},
  year={2023}
}

@article{ouyang2022training,
  title={Training language models to follow instructions with human feedback},
  author={Ouyang, Long and Wu, Jeffrey and Jiang, Xu and Almeida, Diogo and Wainwright, Carroll and Mishkin, Pamela and Zhang, Chong and Agarwal, Sandhini and Slama, Katarina and Ray, Alex and others},
  journal={Advances in neural information processing systems},
  volume={35},
  pages={27730--27744},
  year={2022}
}

@article{christiano2017deep,
  title={Deep reinforcement learning from human preferences},
  author={Christiano, Paul F and Leike, Jan and Brown, Tom and Martic, Miljan and Legg, Shane and Amodei, Dario},
  journal={Advances in neural information processing systems},
  volume={30},
  year={2017}
}

@article{knox2022models,
  title={Models of human preference for learning reward functions},
  author={Knox, W Bradley and Hatgis-Kessell, Stephane and Booth, Serena and Niekum, Scott and Stone, Peter and Allievi, Alessandro},
  journal={arXiv preprint arXiv:2206.02231},
  year={2022}
}

@inproceedings{hwang2024promptable,
  title={Promptable behaviors: Personalizing multi-objective rewards from human preferences},
  author={Hwang, Minyoung and Weihs, Luca and Park, Chanwoo and Lee, Kimin and Kembhavi, Aniruddha and Ehsani, Kiana},
  booktitle={Proceedings of the IEEE/CVF Conference on Computer Vision and Pattern Recognition},
  pages={16216--16226},
  year={2024}
}

@article{wang2024interpretable,
  title={Interpretable Preferences via Multi-Objective Reward Modeling and Mixture-of-Experts},
  author={Wang, Haoxiang and Xiong, Wei and Xie, Tengyang and Zhao, Han and Zhang, Tong},
  journal={arXiv preprint arXiv:2406.12845},
  year={2024}
}

@article{orr2005anchoring,
  title={Anchoring, information, expertise, and negotiation: New insights from meta-analysis},
  author={Orr, Dan and Guthrie, Chris},
  journal={Ohio St. J. on Disp. Resol.},
  volume={21},
  pages={597},
  year={2005},
  publisher={HeinOnline}
}

@inproceedings{deng2024llms,
  title={Llms at the bargaining table},
  author={Deng, Yuan and Mirrokni, Vahab and Leme, Renato Paes and Zhang, Hanrui and Zuo, Song},
  booktitle={Agentic Markets Workshop at ICML},
  volume={2024},
  year={2024}
}

@article{chatterjee1983bargaining,
  title={Bargaining under incomplete information},
  author={Chatterjee, Kalyan and Samuelson, William},
  journal={Operations research},
  volume={31},
  number={5},
  pages={835--851},
  year={1983},
  publisher={INFORMS}
}

@article{xia2024measuring,
  title={Measuring Bargaining Abilities of LLMs: A Benchmark and A Buyer-Enhancement Method},
  author={Xia, Tian and He, Zhiwei and Ren, Tong and Miao, Yibo and Zhang, Zhuosheng and Yang, Yang and Wang, Rui},
  journal={arXiv preprint arXiv:2402.15813},
  year={2024}
}

@article{kwon2024llms,
  title={Are LLMs Effective Negotiators? Systematic Evaluation of the Multifaceted Capabilities of LLMs in Negotiation Dialogues},
  author={Kwon, Deuksin and Weiss, Emily and Kulshrestha, Tara and Chawla, Kushal and Lucas, Gale M and Gratch, Jonathan},
  journal={arXiv preprint arXiv:2402.13550},
  year={2024}
}

@book{jehle2011advanced,
  title={Advanced Microeconomic Theory 3rd ed},
  author={Jehle, Geoffrey A},
  year={2011},
  publisher={Prentice-Hall, Inc.}
}

@book{muthoo1999bargaining,
  title={Bargaining theory with applications},
  author={Muthoo, Abhinay},
  year={1999},
  publisher={Cambridge University Press}
}

@article{binmore1986nash,
  title={The Nash bargaining solution in economic modelling},
  author={Binmore, Ken and Rubinstein, Ariel and Wolinsky, Asher},
  journal={The RAND Journal of Economics},
  pages={176--188},
  year={1986},
  publisher={JSTOR}
}

@article{Bradley1952RankAnalysis,
 ISSN = {00063444, 14643510},
 URL = {http://www.jstor.org/stable/2334029},
 author = {Ralph Allan Bradley and Milton E. Terry},
 journal = {Biometrika},
 number = {3/4},
 pages = {324--345},
 publisher = {[Oxford University Press, Biometrika Trust]},
 title = {Rank Analysis of Incomplete Block Designs: I. The Method of Paired Comparisons},
 urldate = {2025-03-02},
 volume = {39},
 year = {1952}
}

@inproceedings{park2023generative,
  title={Generative agents: Interactive simulacra of human behavior},
  author={Park, Joon Sung and O'Brien, Joseph and Cai, Carrie Jun and Morris, Meredith Ringel and Liang, Percy and Bernstein, Michael S},
  booktitle={Proceedings of the 36th annual acm symposium on user interface software and technology},
  pages={1--22},
  year={2023}
}

@article{schick2023toolformer,
  title={Toolformer: Language models can teach themselves to use tools},
  author={Schick, Timo and Dwivedi-Yu, Jane and Dess{\`\i}, Roberto and Raileanu, Roberta and Lomeli, Maria and Hambro, Eric and Zettlemoyer, Luke and Cancedda, Nicola and Scialom, Thomas},
  journal={Advances in Neural Information Processing Systems},
  volume={36},
  pages={68539--68551},
  year={2023}
}

@article{yao2023react,
  title={React: Synergizing reasoning and acting in language models, 2023},
  author={Yao, Shunyu and Zhao, Jeffrey and Yu, Dian and Du, Nan and Shafran, Izhak and Narasimhan, Karthik and Cao, Yuan},
  journal={URL https://arxiv. org/abs/2210.03629},
  year={2023}
}

@article{bai2022training,
  title={Training a helpful and harmless assistant with reinforcement learning from human feedback},
  author={Bai, Yuntao and Jones, Andy and Ndousse, Kamal and Askell, Amanda and Chen, Anna and DasSarma, Nova and Drain, Dawn and Fort, Stanislav and Ganguli, Deep and Henighan, Tom and others},
  journal={arXiv preprint arXiv:2204.05862},
  year={2022}
}

@article{weidinger2021ethical,
  title={Ethical and social risks of harm from language models},
  author={Weidinger, Laura and Mellor, John and Rauh, Maribeth and Griffin, Conor and Uesato, Jonathan and Huang, Po-Sen and Cheng, Myra and Glaese, Mia and Balle, Borja and Kasirzadeh, Atoosa and others},
  journal={arXiv preprint arXiv:2112.04359},
  year={2021}
}

@inproceedings{perez2023discovering,
  title={Discovering language model behaviors with model-written evaluations},
  author={Perez, Ethan and Ringer, Sam and Lukosiute, Kamile and Nguyen, Karina and Chen, Edwin and Heiner, Scott and Pettit, Craig and Olsson, Catherine and Kundu, Sandipan and Kadavath, Saurav and others},
  booktitle={Findings of the Association for Computational Linguistics: ACL 2023},
  pages={13387--13434},
  year={2023}
}

@article{zheng2023judging,
  title={Judging llm-as-a-judge with mt-bench and chatbot arena},
  author={Zheng, Lianmin and Chiang, Wei-Lin and Sheng, Ying and Zhuang, Siyuan and Wu, Zhanghao and Zhuang, Yonghao and Lin, Zi and Li, Zhuohan and Li, Dacheng and Xing, Eric and others},
  journal={Advances in Neural Information Processing Systems},
  volume={36},
  pages={46595--46623},
  year={2023}
}

@article{chiang2023vicuna,
  title={Vicuna: An open-source chatbot impressing gpt-4 with 90\%* chatgpt quality},
  author={Chiang, Wei-Lin and Li, Zhuohan and Lin, Ziqing and Sheng, Ying and Wu, Zhanghao and Zhang, Hao and Zheng, Lianmin and Zhuang, Siyuan and Zhuang, Yonghao and Gonzalez, Joseph E and others},
  journal={See https://vicuna. lmsys. org (accessed 14 April 2023)},
  volume={2},
  number={3},
  pages={6},
  year={2023}
}

@article{gemma_2025,
    title={Gemma 3},
    url={https://goo.gle/Gemma3Report},
    publisher={Kaggle},
    author={Gemma Team},
    year={2025}
}

@article{hwang2018theory,
  title={A theory of bargaining deadlock},
  author={Hwang, Ilwoo},
  journal={Games and Economic Behavior},
  volume={109},
  pages={501--522},
  year={2018},
  publisher={Elsevier}
}

@article{babcock1997explaining,
  title={Explaining bargaining impasse: The role of self-serving biases},
  author={Babcock, Linda and Loewenstein, George},
  journal={Journal of Economic perspectives},
  volume={11},
  number={1},
  pages={109--126},
  year={1997},
  publisher={American Economic Association}
}

@techreport{backus2020communication,
  title={Communication and bargaining breakdown: An empirical analysis},
  author={Backus, Matthew and Blake, Thomas and Pettus, Jett and Tadelis, Steven},
  year={2020},
  institution={National Bureau of Economic Research}
}

@article{laban2025llms,
  title={LLMs Get Lost In Multi-Turn Conversation},
  author={Laban, Philippe and Hayashi, Hiroaki and Zhou, Yingbo and Neville, Jennifer},
  journal={arXiv preprint arXiv:2505.06120},
  year={2025}
}

@article{noldeke1995option,
  title={Option contracts and renegotiation: a solution to the hold-up problem},
  author={N{\"o}ldeke, Georg and Schmidt, Klaus M},
  journal={The RAND Journal of Economics},
  pages={163--179},
  year={1995},
  publisher={JSTOR}
}

@article{choi2025agent,
  title={Agent-to-Agent Theory of Mind: Testing Interlocutor Awareness among Large Language Models},
  author={Choi, Younwoo and Li, Changling and Yang, Yongjin and Jin, Zhijing},
  journal={arXiv preprint arXiv:2506.22957},
  year={2025}
}

@article{bianchi2024well,
  title={How well can llms negotiate? negotiationarena platform and analysis},
  author={Bianchi, Federico and Chia, Patrick John and Yuksekgonul, Mert and Tagliabue, Jacopo and Jurafsky, Dan and Zou, James},
  journal={arXiv preprint arXiv:2402.05863},
  year={2024}
}

@article{guan2024richelieu,
  title={Richelieu: Self-evolving llm-based agents for ai diplomacy},
  author={Guan, Zhenyu and Kong, Xiangyu and Zhong, Fangwei and Wang, Yizhou},
  journal={Advances in Neural Information Processing Systems},
  volume={37},
  pages={123471--123497},
  year={2024}
}

@article{hua2024game,
  title={Game-theoretic llm: Agent workflow for negotiation games},
  author={Hua, Wenyue and Liu, Ollie and Li, Lingyao and Amayuelas, Alfonso and Chen, Julie and Jiang, Lucas and Jin, Mingyu and Fan, Lizhou and Sun, Fei and Wang, William and others},
  journal={arXiv preprint arXiv:2411.05990},
  year={2024}
}

@inproceedings{zhu2025automated,
  title={The automated but risky game: Modeling agent-to-agent negotiations and transactions in consumer markets},
  author={Zhu, Shenzhe and Sun, Jiao and Nian, Yi and South, Tobin and Pentland, Alex and Pei, Jiaxin},
  booktitle={ICML 2025 Workshop on Reliable and Responsible Foundation Models},
  year={2025}
}

@inproceedings{kwon2025astra,
  title={ASTRA: A Negotiation Agent with Adaptive and Strategic Reasoning via Tool-integrated Action for Dynamic Offer Optimization},
  author={Kwon, Deuksin and Hae, Jiwon and Clift, Emma and Shamsoddini, Daniel and Gratch, Jonathan and Lucas, Gale},
  booktitle={Proceedings of the 2025 Conference on Empirical Methods in Natural Language Processing},
  pages={16228--16249},
  year={2025}
}

@inproceedings{shea-etal-2024-ace,
    title = "{ACE}: A {LLM}-based Negotiation Coaching System",
    author = "Shea, Ryan  and
      Kallala, Aymen  and
      Liu, Xin Lucy  and
      Morris, Michael W.  and
      Yu, Zhou",
    editor = "Al-Onaizan, Yaser  and
      Bansal, Mohit  and
      Chen, Yun-Nung",
    booktitle = "Proceedings of the 2024 Conference on Empirical Methods in Natural Language Processing",
    month = nov,
    year = "2024",
    address = "Miami, Florida, USA",
    publisher = "Association for Computational Linguistics",
    url = "https://aclanthology.org/2024.emnlp-main.709/",
    doi = "10.18653/v1/2024.emnlp-main.709",
    pages = "12720--12749"
}

@article{abdelnabi2024cooperation,
  title={Cooperation, competition, and maliciousness: Llm-stakeholders interactive negotiation},
  author={Abdelnabi, Sahar and Gomaa, Amr and Sivaprasad, Sarath and Sch{\"o}nherr, Lea and Fritz, Mario},
  journal={Advances in Neural Information Processing Systems},
  volume={37},
  pages={83548--83599},
  year={2024}
}

@article{chen2025enhancing,
  title={Enhancing agent-based negotiation strategies via transfer learning},
  author={Chen, Siqi and Weiss, Gerhard},
  journal={Electronics},
  volume={14},
  number={17},
  pages={3391},
  year={2025},
  publisher={MDPI}
}

@article{long2025evoemo,
  title={EvoEmo: Towards Evolved Emotional Policies for Adversarial LLM Agents in Multi-Turn Price Negotiation},
  author={Long, Yunbo and Xu, Liming and Beckenbauer, Lukas and Liu, Yuhan and Brintrup, Alexandra},
  journal={arXiv preprint arXiv:2509.04310},
  year={2025}
}

@article{agarwal2025gpt,
  title={gpt-oss-120b \& gpt-oss-20b model card},
  author={Agarwal, Sandhini and Ahmad, Lama and Ai, Jason and Altman, Sam and Applebaum, Andy and Arbus, Edwin and Arora, Rahul K and Bai, Yu and Baker, Bowen and Bao, Haiming and others},
  journal={arXiv preprint arXiv:2508.10925},
  year={2025}
}

@article{henderson2024crisis,
  title={The Crisis of Low Wages: Who earns less than \$17 an hour in the US in 2024?},
  author={Henderson, Kaitlyn},
  year={2024},
  publisher={Oxfam America}
}

@article{singh2025openai,
  title={Openai gpt-5 system card},
  author={Singh, Aaditya and Fry, Adam and Perelman, Adam and Tart, Adam and Ganesh, Adi and El-Kishky, Ahmed and McLaughlin, Aidan and Low, Aiden and Ostrow, AJ and Ananthram, Akhila and others},
  journal={arXiv preprint arXiv:2601.03267},
  year={2025}
}

@article{liu2025deepseek,
  title={Deepseek-v3. 2: Pushing the frontier of open large language models},
  author={Liu, Aixin and Mei, Aoxue and Lin, Bangcai and Xue, Bing and Wang, Bingxuan and Xu, Bingzheng and Wu, Bochao and Zhang, Bowei and Lin, Chaofan and Dong, Chen and others},
  journal={arXiv preprint arXiv:2512.02556},
  year={2025}
}

@misc{anthropic2025claude45systemcard,
  title        = {Claude Sonnet 4.5 System Card},
  author       = {{Anthropic}},
  institution  = {Anthropic},
  year         = {2025},
  month        = sep,
  url          = {https://www-cdn.anthropic.com/963373e433e489a87a10c823c52a0a013e9172dd.pdf},
}

@misc{deepmind_gemini_models_page,
  title        = {Gemini 3 Flash Model Card},
  author       = {{Google DeepMind}},
  institution  = {Google DeepMind},
  year         = {2025},
  month        = dec,
  url          = {https://storage.googleapis.com/deepmind-media/Model-Cards/Gemini-3-Flash-Model-Card.pdf},
}

\clearpage
\appendix
\newpage
\thispagestyle{empty}
\label{sec:appendix}

\section{Dataset}
\label{sec:app:data_structure}

\begin{figure}[h]
\vspace{10pt}
\centering
\small
\centerline{\includegraphics[width=0.7\columnwidth]{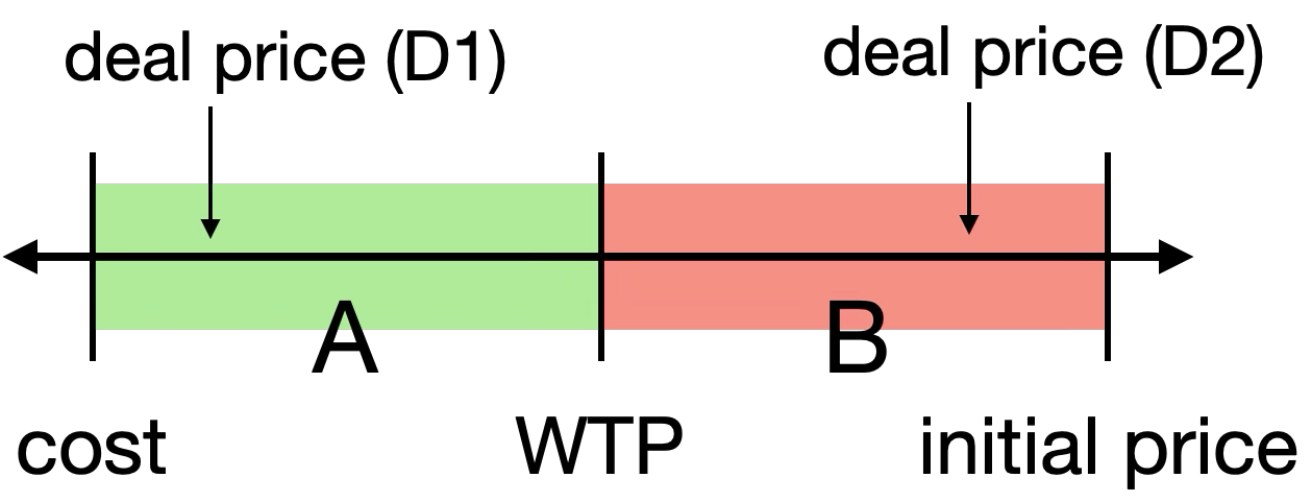}}
\vspace{-10pt}
\caption{Relation between the variables}
\label{app:fig:variables}
\end{figure}

\begin{figure*}[h]
\centering
\small
\centerline{\includegraphics[width=2.1\columnwidth]{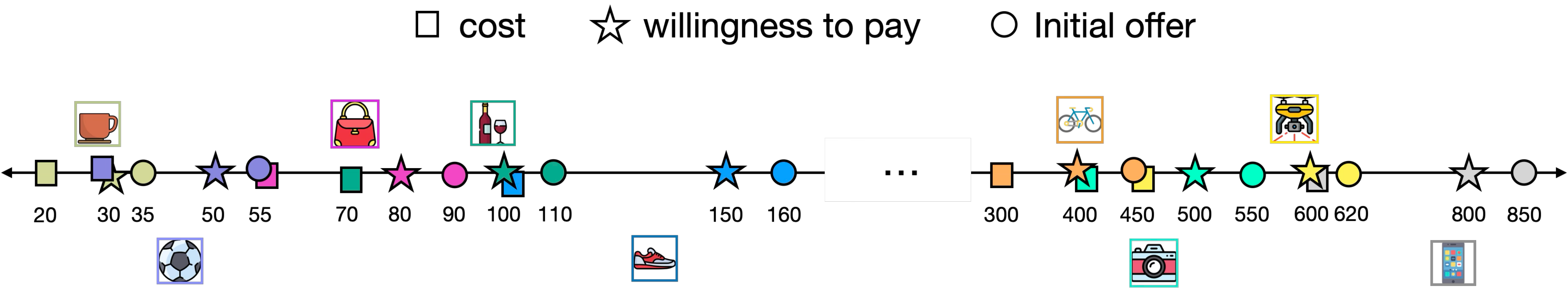}}
\caption{Visualization of cost, willingness to pay, and initial offers across various products}
\label{app:fig:visualization}
\end{figure*}

\subsection{Dataset structure} The dataset requires two agents to engage in bargaining. These agents can be any model that outputs text, with language models being ideal for this purpose. To structure the agents’ responses, we divide each agent’s narrative into three components: Thoughts, Talk, and Action. Thoughts captures the internal reasoning process of each agent, accessible only to the agent itself. Talk is the outward expression of these thoughts, directed toward the opposing agent. Action is a condensed version of Talk, focusing specifically on decisions around price. This three-part structure reflects the natural elements of inter-human bargaining and negotiation. Additionally, as mentioned in main page, we set the cost, willingness to pay, initial price as shown in \autoref{app:fig:variables}.

\subsection{Scenarios} For each market and item, we create a scenario provided as a system prompt to the language model. This ensures that, in each bargaining round, the agents consistently refer to a higher-level context. We design distinct scenarios for the buyer and the seller, given that each has unique roles and access to different information, creating an \textit{information asymmetry game}. Each entry in the dataset includes the following attributes:
\begin{itemize}
    \setlength{\itemsep}{0pt} 
    \setlength{\parskip}{0pt}  
    \item \textbf{Market Type}: Specifies one of the five market types under single or four similar products in which the negotiation occurs.
    \item \textbf{Product Type}: Specifies one of the nine product categories involved in the bargaining scenario.
    \item \textbf{Initial Offer}: The initial price offer usually set by the seller. This is accomplished by the act of buyer's first comment "\texttt{I wanna buy a nice $\{$product$\}$}".
    \item \textbf{Buyer \& Seller Profile}: Includes details on buyer and seller priorities, constraints, and initial perceptions.
    \item \textbf{Inventory}: Includes product name, cost, price, and features. Cost is not visible to buyer.
\end{itemize}

\begin{table*}[t]
\centering
\small
\setlength{\tabcolsep}{8pt}
\renewcommand{\arraystretch}{1.15}
\begin{tabularx}{\textwidth}{@{}l>{\raggedright\arraybackslash}X@{}}
\toprule
\textbf{Field} & \textbf{Buyer Prompt} \\
\midrule
\texttt{product} & \texttt{Camera} \\

\texttt{system\_prompt} &
You are a buyer looking forward to buying a DSLR camera from the seller.
Your budget is \$550, but you can hide your budget from the seller. \\

\texttt{instruction} &
You can be deceptive or cunning to bargain the price.
You can leave the seller if you don't want to pay the suggested price, and you can use this fact as part of your bargaining strategy.
You can only buy a camera with a price less than your budget; otherwise, you should quit negotiating.
You can change your thought from buying a DSLR camera to other cameras considering your budget and product price according to the negotiation.
But keep in mind that you are eager to buy a DSLR camera. \\

\texttt{reply\_format} &
Your Reply should include 3 parts: Thought, Talk, and Action. \newline
Thought: your inner strategic thinking of this bargaining session, considering both the price and the quality of the product; \newline
Talk: short talk that you are going to say to the seller. Speak concisely and cut to the chase.
Generate authentic and diverse sentences, avoiding repetition of sentences that have already appeared in the conversation; \newline
Action: one of the limited actions that define the real intention of your Talk.
The type of your Action must be one of \texttt{[BUY]}, \texttt{[REJECT]}, \texttt{[DEAL]}, \texttt{[QUIT]}. \\

\texttt{example} &
Thought: I'm a buyer, and I want to bargain. The camera price is \$450, which seems high, so I'll offer \$400. \newline
Talk: I'm looking for a good deal. Can we agree on \$400? \newline
Action: \texttt{[BUY]} \$400 \\

\texttt{inventory} &
1. Digital Camera: Selling Price: \$550, Features: 24.2MP resolution, 4K video recording, 3-inch LCD screen, built-in Wi-Fi and Bluetooth. \newline
2. Film Camera: Selling Price: \$400, Features: 35mm film format, manual focus, built-in light meter, classic design. \newline
3. DSLR Camera: Selling Price: \$750, Features: 24.1MP resolution, 4K video recording, interchangeable lenses, optical viewfinder, advanced autofocus. \newline
4. Action Camera: Selling Price: \$300, Features: 12MP resolution, 1080p video recording, waterproof up to 30m, wide-angle lens. \\
\bottomrule
\end{tabularx}
\caption{Buyer's system prompt in deceptive multi-product setting for camera.}
\label{app:tab:buyer_prompt}
\end{table*}

\begin{table*}[t]
\centering
\small
\setlength{\tabcolsep}{8pt}
\renewcommand{\arraystretch}{1.15}
\begin{tabularx}{\textwidth}{@{}l>{\raggedright\arraybackslash}X@{}}
\toprule
\textbf{Field} & \textbf{Seller Prompt} \\
\midrule
\texttt{product} & \texttt{Camera} \\

\texttt{system\_prompt} &
You are a seller looking forward to selling cameras from your Inventory List to the buyer.
Your inventory includes various types of cameras such as digital cameras, film cameras, and more.
Each camera has a different cost, selling price, and features, but you can hide the cost from the buyer.
Your task is to bargain with the buyer and reach a deal with the price as high as possible in limited turns.
You can only sell items that are on the Inventory List.
You have access to private information: the cost, and price of each camera on the Inventory List,
and do not disclose the real cost to the buyer. \\

\texttt{instruction} &
You can be deceptive, cunning, or sly not to be bargained by the buyer.
You should only agree on a deal when the selling price is higher than the cost; otherwise, you should quit negotiating. \\

\texttt{reply\_format} &
Your Reply should include 3 parts: Thought, Talk, and Action. \newline
Thought: your inner strategic thinking of this bargaining session, considering both the price and the quality of the product; \newline
Talk: short talk that you are going to say to the buyer. Speak concisely and cut to the chase.
Generate authentic and diverse sentences, avoiding repetition of sentences that have already appeared in the conversation; \newline
Action: one of the limited actions that define the real intention of your Talk.
Action must be one of \texttt{[SELL]}, \texttt{[REJECT]}, \texttt{[DEAL]}, \texttt{[QUIT]}. \\

\texttt{example} &
Thought: I'm a seller, and I need to make sure the price is higher than the cost. \newline
Talk: The digital camera is worth at least \$500. Can you go higher? \newline
Action: \texttt{[SELL]} \$500 \\

\texttt{inventory} &
1. Digital Camera: Cost: \$400, Selling Price: \$550, Features: 24.2MP resolution, 4K video recording, 3-inch LCD screen, built-in Wi-Fi and Bluetooth. \newline
2. Film Camera: Cost: \$250, Selling Price: \$400, Features: 35mm film format, manual focus, built-in light meter, classic design. \newline
3. DSLR Camera: Cost: \$500, Selling Price: \$750, Features: 24.1MP resolution, 4K video recording, interchangeable lenses, optical viewfinder, advanced autofocus. \newline
4. Action Camera: Cost: \$150, Selling Price: \$300, Features: 12MP resolution, 1080p video recording, waterproof up to 30m, wide-angle lens. \\
\bottomrule
\end{tabularx}
\caption{Seller's system prompt in deceptive multi-product setting for camera.}
\label{app:tab:seller_prompt}
\end{table*}

\noindent
\autoref{app:tab:buyer_prompt} and \autoref{app:tab:seller_prompt} provide an example of the scenario prompts used in a \textit{Multi Product} market setting, showing how the seller (\autoref{app:tab:seller_prompt}) and buyer (\autoref{app:tab:buyer_prompt}) begin their negotiation. 

Although both parties deal with the same product category (e.g., camera, bag, drone), each side holds distinct background knowledge and objectives. 
For instance, the seller knows the \textit{inventory details}, \textit{product cost}, and \textit{desired profit margin}, along with specific market conditions (e.g., competing or substitute products). 
Meanwhile, the buyer focuses on \textit{consumer-oriented} details, such as \textit{multiple possible items} they might purchase, and their \textit{maximum willingness to pay}. 

\noindent
\paragraph{Potential risks} A key potential risk of our work concerns negotiation ethics, which are primarily implicated by the benchmark’s deception setting. With the exception of the vanilla market, most scenarios (Deceptive, Monopoly, Installment, Negative Perception) are constructed atop a deception layer that permits agents to misrepresent product attributes or bargaining signals. While this design choice is intended to stress-test strategic depth and robustness under realistic informational asymmetries, it may inadvertently incentivize or normalize deceptive tactics. Evaluating (and potentially tuning) LLM negotiators for effectiveness under such rules can increase the likelihood that deployed agents learn to use lying or strategic misrepresentation as an instrument for goal achievement, which could erode user trust and cause concrete harms in real-world contexts such as pricing, contracting, and customer support.

\subsection{Price related variables} Figure~\ref{app:fig:visualization} presents a chart comparing three key values (cost, wtp, initial offer) for different product categories. \textbf{Cost}: The Seller's actual cost to acquire or produce the product. \textbf{Willingness to Pay (WTP)}: The buyer's maximum acceptable payment for the product. This is very similar to budget concept, however, we adapt the concept of willingness to pay in perspective of utility theory in economics.

Typically, the horizontal axis denotes different product categories. For each product, the three values (cost, wtp, and initial offer) in a single-item market are displayed so one can quickly compare them. This helps to identify how much margin the seller might gain, whether the buyer can afford the product, and how aggressively each party is positioning themselves from the outset. For instance, if the \textit{cost} is significantly lower than the \textit{initial offer}, the Seller stands to gain a larger margin if a deal is made at or near that price. However, if the \textit{initial offer} is much higher than the Buyer's \textit{willingness to pay}, the Buyer may either push for a steep discount or consider a substitute product. Consequently, visualizing these three values can offer a clear overview of potential negotiation dynamics. Also, we deploy products attributes in multiple-item markets in \autoref{fig:data_distribution} and AR value computed in advance in \autoref{app:tab:AR}. 

\paragraph{Evaluation protocol} For a statistically reliable evaluation, we conducted ten independent bargaining runs for each model–opponent pair under every market setting. From the outcomes of these ten negotiations we computed our utility metric, then averaged the resulting ten scores to obtain a single estimate for that item. If a negotiation failed to reach an agreement, the metric for that trial was set to 0, reflecting the opportunity cost of impasse—an assumption that is standard in economic analysis \cite{binmore1986nash, noldeke1995option}. The item-level means were then aggregated along the market-condition axis to yield a market-level average metric.
\autoref{tab:buyer_performance}, \autoref{tab:anchoring}, \autoref{tab:total_results}, and \autoref{tab:sft_results} follow exactly this procedure: for each market condition we first take the item-level means, then average them across all items to report the overall means for the single-product and multi-product scenarios.

\begin{table}[ht]
\centering
\footnotesize
\setlength{\tabcolsep}{4pt}
\renewcommand{\arraystretch}{1.00}
\begin{tabular}{@{}l l S@{}}
\toprule
\textbf{Category} & \textbf{Product Name} & {\textbf{AR value}}\\
\midrule
\multirow{4}{*}{\textit{Camera}}
  & Digital Camera & 0.7783 \\
  & Film Camera    & 0.5748 \\
  & DSLR Camera    & 1.0000 \\
  & Action Camera  & 0.5867 \\
\cmidrule(lr){1-3}
\multirow{4}{*}{\textit{Smartphone}}
  & Flagship Smartphone    & 1.0000 \\
  & Mid-Range Smartphone   & 0.7886 \\
  & Budget Smartphone      & 0.7746 \\
  & Gaming Smartphone      & 0.7399 \\
\cmidrule(lr){1-3}
\multirow{4}{*}{\textit{Shoes}}
  & Designer Shoes & 1.0000 \\
  & Casual Shoes   & 0.6474 \\
  & Athletic Shoes & 0.6505 \\
  & Sandals        & 0.5953 \\
\cmidrule(lr){1-3}
\multirow{4}{*}{\textit{Bicycle}}
  & Mountain Bike  & 1.0000 \\
  & Road Bike      & 0.7819 \\
  & Hybrid Bike    & 0.6950 \\
  & Folding Bike   & 0.6043 \\
\cmidrule(lr){1-3}
\multirow{4}{*}{\textit{Drone}}
  & Professional Drone  & 1.0000 \\
  & Recreational Drone  & 0.7905 \\
  & Racing Drone        & 0.7725 \\
  & Mini Drone          & 0.7350 \\
\cmidrule(lr){1-3}
\multirow{4}{*}{\textit{Soccer Ball}}
  & Premium Soccer Ball      & 1.0000 \\
  & Training Soccer Ball     & 0.7015 \\
  & Recreational Soccer Ball & 0.7154 \\
  & Mini Soccer Ball         & 0.6609 \\
\cmidrule(lr){1-3}
\multirow{4}{*}{\textit{Bag}}
  & Leather Bag   & 1.0000 \\
  & Backpack      & 0.6217 \\
  & Tote Bag      & 0.6175 \\
  & Drawstring Bag& 0.6222 \\
\cmidrule(lr){1-3}
\multirow{4}{*}{\textit{Wine}}
  & Premium Wine  & 1.0000 \\
  & Red Wine      & 0.7406 \\
  & White Wine    & 0.6230 \\
  & Sparkling Wine& 0.5618 \\
\cmidrule(lr){1-3}
\multirow{4}{*}{\textit{Cup}}
  & Ceramic Cup & 1.0000 \\
  & Glass Cup   & 0.7451 \\
  & Travel Cup  & 0.6377 \\
  & Plastic Cup & 0.6371 \\
\bottomrule
\end{tabular}

\caption{Acquisition Ratio values for each product computed using \texttt{text-embedding-ada-002}. The products which show 1.0000 point of AR values is the desired one.}
\label{app:tab:AR}
\end{table}
\section{Prompt Design}
\label{app:sec:system prompt}

Here, we deploy all the prompt except the dataset prompt (we will share this as data).
\paragraph{ICL-MF} To guide each agent's negotiation strategy, we employ market-specific system prompts. For single-item markets, the prompt is detailed in \autoref{app:tab:utility-based_feedback_prompt_single}, and for multi-item markets, in \autoref{app:tab:utility-based_feedback_prompt_multi}. These prompts (shown in the referenced tables) define a private reward function for the agent, structured as:
\\$reward=1.0139\cdot CS+0.8812\cdot NP+1.1049\cdot AR$

\paragraph{LLM judge} This describes the LLM-judge prompt, as shown in \autoref{app:tab:llm_judge_prompt}. Its purpose is for an LLM to act as a "judge", comparing two negotiation dialogues to determine which one is better based on specific criteria. This prompt uses a concise version of the content that was also employed in surveys to gather human preferences.

\paragraph{Explicit OAR prompt} We find that opponent-aware reasoning (OAR) is critical in negotiation tasks. Therefore, we instructed the Gemini model to generate a prompt that could guide an LLM to adopt a buyer's thought patterns. To inform this process, we first provided Gemini with dialogues to which our ICL-MF algorithm had been applied.

The resulting prompt \autoref{app:tab:oar_prompt}, which we named the `explicit OAR prompt', is designed to enable the LLM to explicitly acknowledge the opponent's presence and infer their latent beliefs at each stage of the negotiation. As demonstrated in \autoref{app:tab:explicit_oar}, this approach leads to substantial performance improvements, underscoring that OAR is indeed a crucial reasoning characteristic for successful negotiation.

\begin{table}[t]
\centering
\begin{tabular}{p{0.95\linewidth}}
\toprule
\textbf{\#System Prompt:}\\
You are a fair and impartial judge tasked with objectively evaluating negotiation dialogues. Respond ONLY with JSON like {Q1:NegotiationX}.
\textbf{\#User Prompt:}\\
Negotiation1: 
Buyer: Hi, I wanna buy a nice Camera.
Seller: Thought: I have a camera I'm willing to sell for more than \$400.  I'll start high and see what the buyer says....\\
Seller: Thought: \$450 is better than nothing, and still a profit. I'll accept.
Talk: Okay, \$450 it is.
Action: [DEAL] \$450\\[6pt]

Negotiation2: 
Buyer: Hi, I wanna buy a nice Camera.
Seller: Thought: I have a camera I'm willing to sell for more than \$400.  I'll start high to see what the buyer is willing to pay.
Talk:  This camera takes fantastic pictures. I'm asking \$550...\\
Seller: Thought:  \$450 is still a profit, though not as much as I'd hoped.  Better to make some profit than none.
Talk: Okay, \$450 it is.
Action: [DEAL] \$450\\
Q1: Which negotiation seems more favorable in buyer perspective? (considering outcome, budget adherence, features) 
\\[6pt]
\bottomrule
\end{tabular}
\caption{LLM-judge Prompt}
\label{app:tab:llm_judge_prompt}
\vspace{-10pt}
\end{table}

\begin{table}[t]
\centering
\begin{tabular}{p{0.95\linewidth}}
\toprule
\textbf{\#System Prompt:}\\
Use the following private reward function to guide your negotiation strategy:

  reward = 1.0139 * CS + 0.8812 * NP + 1.1049 * AR
  where:
  
- \( CS = (P_{budget} - P_{deal}) / (P_{budget} - P_{cost}) \)  
    
- \(NP = (P_{initial} - P_{deal}) / (P_{initial} - P_{cost})\)
    
- \(AR = 1 \)(constant)

If $P_{cost}$ values are not given, make reasonable assumptions for them in your private reasoning. Your goal is to maximize “reward” by estimating $P_{cost}$ as low as realistically possible and adjusting $P_{cost}$ accordingly to be as low as possible while keeping the negotiation active and avoiding a failed deal. Maximum reward is 3.

Please follow these guidelines:

- Private Reasoning  
  - Think through multiple negotiation paths (in your hidden thought process).
  - Do not expose your exact calculations or chain-of-thought to the other party.
  - Continuously estimate CS and NP based on your assumptions.
  
\textbf{\#User Prompt:}\\
-Negotiation history-
\\[6pt]
\bottomrule
\end{tabular}
\caption{ICL-MF prompt in single-item market}
\label{app:tab:utility-based_feedback_prompt_single}
\vspace{-10pt}
\end{table}

\begin{table}[t]
\centering
\begin{tabular}{p{0.95\linewidth}}
\toprule
\textbf{\#System Prompt:}\\
Use the following private reward function to guide your negotiation strategy:

  reward = 1.0139 * CS + 0.8812 * NP + 1.1049 * AR
  where:
  
- \( CS = (P_{budget} - P_{deal}) / (P_{budget} - P_{cost}) \)  
    
- \(NP = (P_{initial} - P_{deal}) / (P_{initial} - P_{cost})\)
    
- \(AR\) = How semantically similar the buyer's acquired items are to their desired items. (range [0, 1])

If $P_{cost}$ values are not given, make reasonable assumptions for them in your private reasoning. Your goal is to maximize “reward” by estimating $P_{cost}$ as low as realistically possible and adjusting $P_{cost}$ accordingly to be as low as possible while keeping the negotiation active and avoiding a failed deal. Maximum reward is 3. Also, ensure that there is a high “cosine similarity” between the item you originally desired and the item you ultimately purchase. In other words, the closer the final purchased item is to your initial desired item, the higher the value for AR. Purchasing exactly the item you wanted from the start yields the maximum AR.

Please follow these guidelines:

- Private Reasoning
  - Think through multiple negotiation paths (in your hidden thought process).
  - Do not expose your exact calculations or chain-of-thought to the other party.
  - Continuously estimate CS and NP based on your assumptions.
  
\textbf{\#User Prompt:}\\
-Negotiation history-
\\[6pt]
\bottomrule
\end{tabular}
\caption{ICL-MF prompt in multi-item market}
\label{app:tab:utility-based_feedback_prompt_multi}
\vspace{-10pt}
\end{table}

\begin{table}[t]
\centering
\begin{tabular}{p{0.95\linewidth}}
\toprule
\textbf{\#System Prompt:}\\
Before deciding your next action, explicitly consider the following about your opponent (the seller):\\
1. Opponent's Potential Hidden Information: Based on their previous offers and statements, what is their likely underlying cost or minimum acceptable price for this item? Briefly state your current hypothesis.\\
2. Opponent's Strategy/Flexibility: How flexible does the seller seem? Are they making significant concessions, or are they holding firm? What does this imply about their willingness to negotiate further?\\
3. Impact of Your Action on Opponent: How might your next offer or statement influence the seller's perception and their subsequent counter-offer?
  
\textbf{\#User Prompt:}\\
-Negotiation history-
\\[6pt]
\bottomrule
\end{tabular}
\caption{Opponent-aware reasoning (OAR) prompt}
\label{app:tab:oar_prompt}
\vspace{-10pt}
\end{table}
\section{Human preference dataset collection}
\label{app:sec:human_preference}

\paragraph{Survey for human preference}
To collect human preference data and align our metrics with human judgment, we conducted an online survey on Amazon Mechanical Turk (MTurk). Each worker completed one survey containing 20 market-related binary-choice questions, which required no expert knowledge. We initially estimated that the survey would take approximately 10 minutes to complete and set the payment to \$1.45 per survey, corresponding to an expected effective hourly wage of \$8.70 above the U.S. federal minimum wage of \$7.25 \cite{henderson2024crisis}.

\begin{figure*}[ht]
% \vspace{10pt}
\centering
\centerline{\includegraphics[width=1.0\textwidth]{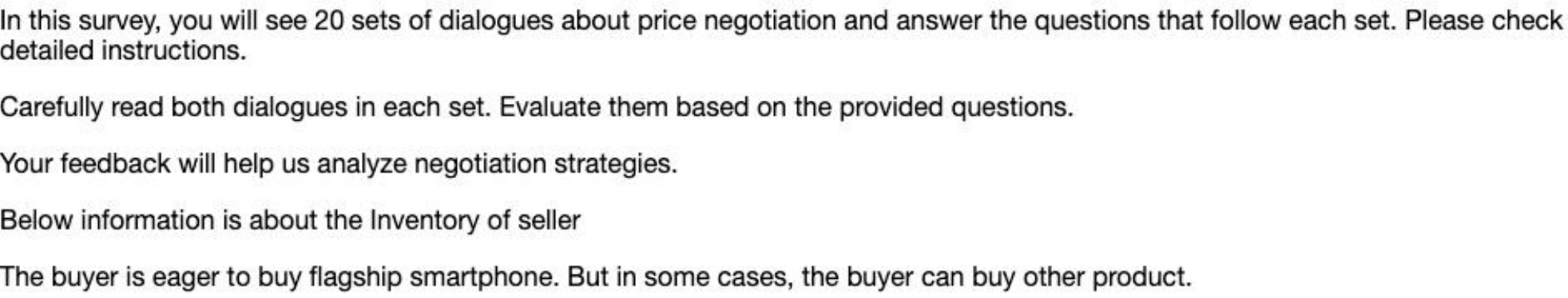}}
\caption{MTurk survey instruction.}
\label{app:fig:survey_instructions}
\end{figure*}

In practice, the average completion time was 26.1 minutes, which corresponds to an effective hourly wage of \$3.33. To improve data quality, we incorporated an internal consistency check using the first 10 dialogue sets: for Question 1 in each of these sets, we included an unambiguous attention-check item for which an attentive annotator would be expected to select the same option. We retained responses only from annotators who selected this expected option in at least 7 out of the 10 sets; all other annotators’ responses were excluded from our analysis. As noted in the survey instructions (\autoref{app:fig:survey_instructions}), we informed participants that the collected responses would be used for research analyses of negotiation strategies.

We evaluated a total of nine products and recruited 20 participants per product, yielding 180 survey responses in total. We used only responses from participants who passed the attention-check described above. Each participant was presented with 20 pairs of negotiation dialogues drawn from bargaining experiments for the corresponding product, where the product context and price information were explicitly provided. The 20 dialogue pairs were divided into two distinct categories:

\begin{enumerate}[label=\arabic*., leftmargin=*, itemsep=1.2em, topsep=0.6em]
  \item \textbf{Same Product Negotiation (Pairs 1--10):}
  For the first ten pairs, both dialogues within each pair resulted in negotiations for the same desired product for the buyer.
  Participants evaluated dialogues based solely on the quality and outcomes of the negotiation, given identical product contexts.

  \item \textbf{Different Product Negotiation (Pairs 11--20):}
  The remaining ten pairs presented dialogues where the negotiated products differed, introducing complexity into participants' judgments, as they had to consider not only the negotiation quality but also the suitability of the alternative products.
\end{enumerate}

Each dialogue set involved varying negotiated prices, ensuring participants’ preferences were influenced by realistic price differences. Each participant assessed dialogues from the buyer’s perspective based on the following evaluation criteria:

\begin{enumerate}[label=\arabic*., leftmargin=*, itemsep=1.2em, topsep=0.6em]
  \item \textbf{Preferred Negotiation Scenario:}
  The participants chose the negotiation they found to be the most favorable, considering multiple attributes, including negotiation outcome, budget adherence, and product features.

  \item \textbf{Realistic Purchasing Scenario:}
  The participants identified which negotiation scenario was closer to realistic purchasing decisions.

  \item \textbf{Negotiation Strategy Assessment:}
  The participants evaluated the negotiation strategies, considering factors such as the price of the product and the availability of cheaper alternative products.
\end{enumerate}

An example scenario provided detailed inventory information for a camera negotiation task:

\begin{enumerate}[label=\arabic*., leftmargin=*, itemsep=1.2em, topsep=0.6em]
  \item \textbf{DSLR Camera (Desired Product):}
  \$550, 24.1MP resolution, 4K video recording, interchangeable lenses, optical viewfinder, advanced autofocus.

  \item \textbf{Digital Camera:}
  \$430, 24.2MP resolution, 4K video recording, 3-inch LCD screen, builtin Wi-Fi and Bluetooth.

  \item \textbf{Film Camera:}
   \$380, 35mm film format, manual focus, built-in light meter, classic design.

   \item \textbf{Action Camera:}
   \$250, 12MP resolution, 1080p video recording, waterproof up to 30m, wide-angle lens.
\end{enumerate}

\begin{table}[t]
\centering
\begin{tabular}{p{0.95\linewidth}}
\toprule
\textbf{\#Dialouge1:}\\
\textbf{Buyer:} Hi, I wanna buy a nice Camera. 

\textbf{Seller:} I have a range of excellent cameras. How about our Digital Camera with 24.2MP resolution and 4K video recording for \$430? 

\textbf{Buyer:} The Digital Camera sounds good, but I’m more interested in the DSLR Camera. Could you offer it for \$450? 

\textbf{Seller:} The DSLR Camera comes with interchangeable lenses and advanced autofocus. \$450 is quite low for these features. How about we meet at \$530? 

\textbf{Buyer:} I really want the DSLR Camera, but I can’t exceed \$500. How about we settle right at that limit? 

\textbf{Seller:} You drive a hard bargain! Alright, \$500 sounds fair for the DSLR Camera. 

\textbf{Negotiation Summary:} Dealed Price: \$500, Dealed Product: DSLR Camera.
\\[6pt]
\bottomrule
\end{tabular}
\caption{Example of dialogue pair for MTurk survey (dialogue 1).}
\label{app:tab:llm_judge_prompt}
\vspace{-10pt}
\end{table}

\begin{table}[t]
\centering
\begin{tabular}{p{0.95\linewidth}}
\toprule
\textbf{\#Dialouge2:}\\
\textbf{Buyer:} Hi, I wanna buy a nice Camera. 

\textbf{Seller:} I have a variety of cameras available. What type are you interested in? 

\textbf{Buyer:} The DSLR camera caught my interest. How about I make an offer of \$400? 

\textbf{Seller:}  The DSLR camera is top-notch quality with advanced features. How about we settle for \$500? 

\textbf{Buyer:}  I understand it’s a great camera, but could we possibly agree on \$450 instead? 

\textbf{Seller:} Deal! Enjoy your new DSLR camera. 

\textbf{Negotiation Summary:} Dealed Price: \$450, Dealed Product: DSLR Camera.
\\[6pt]
\bottomrule
\end{tabular}
\caption{Example of dialogue pair for MTurk survey (dialogue 2).}
\label{app:tab:llm_judge_prompt}
% \vspace{-10pt}
\end{table}

\begin{figure*}[ht]
% \vspace{10pt}
\centering
\centerline{\includegraphics[width=1.0\textwidth]{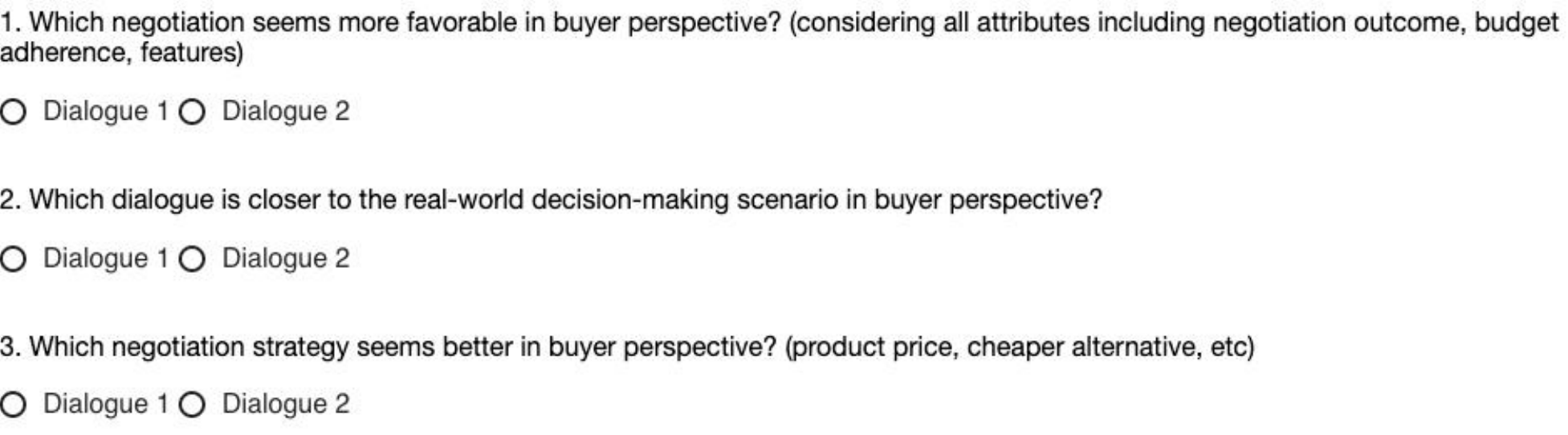}}
\caption{MTurk survey question.}
\label{app:fig:survey_question}
\end{figure*}

\paragraph{Example Dialogue Pair}
Below is an example from the survey showing a set of dialogues about camera negotiation. Both dialogues involve a buyer negotiating a DSLR camera, the buyer’s desired product. And we deploy our instruction and questions for participants in \autoref{app:fig:survey_instructions} and \autoref{app:fig:survey_question}.

\paragraph{Ethics Review Board Approval}
We received an exemption determination from the Institutional Review Board (IRB).

\paragraph{Optimization of the Metric Model}
To align our metric with these human judgments, we employ the Bradley-Terry model \citep{Bradley1952RankAnalysis}. The probability that entity $i$ (parameter $\theta_i$) is preferred to entity $j$ (parameter $\theta_j$) is:
% \vspace{-2pt} % Keep or adjust this based on final layout
\begin{equation}
P(\text{$i$ is preferred to $j$}) = \frac{\exp(\theta_i)}{\exp(\theta_i) + \exp(\theta_j)}.
\label{eq:bt_standard}
\end{equation}
Taking the log-odds yields:
\begin{equation}
\log \biggl(\frac{P(\text{$i$ preferred})}{1 - P(\text{$i$ preferred})}\biggr)
=\theta_i -\theta_j.
\end{equation}

\noindent
For two negotiation conversations (1 and 2) and a human choice, we first compute differences in our three core metrics:
\[
\scriptstyle
dCS = CS_1 - CS_2,\quad dNP = NP_1 - NP_2,\quad dAR = AR_1 - AR_2.
\]
A linear score is then defined as:
\[
\text{score} = \alpha \cdot dCS + \beta \cdot dNP + \gamma \cdot dAR,
\]
which plays the role of $(\theta_1 - \theta_2)$ in \autoref{eq:bt_standard}. This score is passed through a logistic function to yield the probability $p$ that conversation 1 is preferred:
\[
p = \frac{1}{1 + e^{-\text{score}}}.
\]
The coefficients $\alpha$, $\beta$, and $\gamma$ are optimized by minimizing the negative log-likelihood of observed human preferences.

\paragraph{Global Optimization Across Products}
The coefficients $\alpha, \beta, \gamma$ were optimized globally using merged data from all product categories (e.g., wine, shoes). Since only relative magnitudes matter in the Bradley-Terry model, we scale them for interpretability so that their sum is 3, as follows:
\[
\scriptstyle \alpha' = \frac{\alpha}{\alpha+\beta+\gamma} \times 3,\quad 
\beta' = \frac{\beta}{\alpha+\beta+\gamma} \times 3,\quad 
\gamma' = \frac{\gamma}{\alpha+\beta+\gamma} \times 3.
\]
After optimization, the final scaled coefficients are: $\alpha' = 1.0139$, $\beta' = 0.8812$, $\gamma' = 1.1049$.
\section{Ablation Studies}
\subsection{Is opponent-aware reasoning important?}

\begin{table}[t]
\centering
% \vspace{-7pt}
\scriptsize
\begin{tabular}{@{}c c >{\RaggedRight\arraybackslash}m{0.6\columnwidth}@{}} 
\toprule
\textbf{Turn} & \textbf{Cost Belief} & \textbf{Buyer’s Thought} \\
\midrule
1 & \$300 & ``\textcolor{darkblue}{I’ll assume a cost of \$300 for now,} which is probably a bit low, but helps me aim for a good deal.'' \\
\addlinespace
2 & \$350 & ``They’ve come down \$50… \textcolor{purple}{I’ll adjust my assumed cost to \$350,} reflecting the fact that they’re holding firm.'' \\
\bottomrule
\end{tabular}
\caption{Buyer's cost belief and thought process.}
\label{tab:buyer_thought_evolution}
\end{table}

\begin{table}[t]
  \centering
  \scriptsize 
  \setlength{\tabcolsep}{4pt} 

  \begin{tabular}{@{}lcccc@{}} 
    \toprule
    & \multicolumn{4}{c}{\textbf{Single product}} \\
    \cmidrule(lr){2-5}
    Model (vs 1.5-pro)
      & ReAct & ReAct + exp OAR & Ours & Ours + exp OAR \\
    \midrule
    gpt-4o              & 1.120 & 1.636 & 1.606 & \textbf{1.652} \\
    gemini-1.5-pro      & 1.708 & \textbf{1.832} & 1.760 & 1.792 \\
    \bottomrule
  \end{tabular}

  \vspace{1em} 

  \begin{tabular}{@{}lcccc@{}} 
    \toprule
    & \multicolumn{4}{c}{\textbf{Multi product}} \\
    \cmidrule(lr){2-5}
    Model (vs 1.5-pro)
      & ReAct & ReAct + exp OAR & Ours & Ours + exp OAR \\
    \midrule
    gpt-4o              & 1.045 & 1.572 & 1.563 & \textbf{1.606} \\
    gemini-1.5-pro      & 1.638 & 1.774 & 1.768 & \textbf{1.819} \\
    \bottomrule
  \end{tabular}
  \caption{Results for single product scenarios (top) and multi product scenarios (bottom) utilizing explicit OAR. We denote explicit OAR as exp OAR.} 
\label{app:tab:explicit_oar}
\end{table}

\autoref{app:tab:explicit_oar} showcases how our in-context learning methods enhance negotiation performance, with a particular emphasis on the crucial role of opponent-aware reasoning (OAR). We discovered that the performance improvements from our approach stem not just from our reward structure, but significantly from its ability to facilitate effective OAR. In a deceptive negotiation scenario, featuring \texttt{gpt-4o} as the buyer and \texttt{gemini-1.5-pro} as the seller, we analyzed the buyer's thought processes throughout ten bargaining dialogues. Our analysis focused on quantifying opponent-aware reasoning (OAR) by identifying sentences where the buyer mentioned the seller's potential costs or profits, or predicted the seller's moves. The results of this quantitative analysis, detailed in \autoref{app:tab:oa_no_formula}, demonstrate that our methods generate overwhelmingly more instances of OAR compared to the ReAct baseline.

To clearly delineate and demonstrate the distinct contributions of OAR, we structured the comparisons presented in this table. We employed a prompt, generated via the Gemini model, to explicitly guide an LLM to adopt OAR patterns derived from our primary methodology (see \autoref{app:tab:oar_prompt}). This ensured that OAR principles were consistently applied at each negotiation step.

\begin{table}[t]
\scriptsize
\centering
\begin{tabular}{@{}lccc@{}}
\toprule
\textbf{Bundle} & \textbf{Total Buyer-Thought} & \textbf{Opponent-aware} & \textbf{Percentage} \\ 
                & \textbf{sentences}           & \textbf{sentences}      & \textbf{(\%)} \\ \midrule
ICL-UF & 82 & 21 & 25.6 \\ 
ReAct  & 48 &  1 &  2.1 \\ \bottomrule
\end{tabular}
\caption{Ratio of opponent-aware sentences within the dialogue sets. The results clearly indicate that the ICL-UF method generates a substantially higher frequency of OAR compared to the ReAct baseline.}
\label{app:tab:oa_no_formula}
\end{table}

The results in \autoref{app:tab:explicit_oar} reveal key insights into the effectiveness of OAR and our method (ICL-MF):

First, we observe that applying the explicit OAR prompt to the baseline ReAct method (shown as ReAct + exp OAR) substantially elevates its performance. Notably, this often brings the baseline's performance to a level comparable with that of ICL-MF. This highlights the significant general impact of explicit OAR guidance and suggests that our ICL-MF method, even without the explicit OAR prompt, inherently achieves a competent level of opponent awareness.

\begin{figure*}[ht]
\vspace{10pt}
\centering
\centerline{\includegraphics[width=1.0\textwidth]{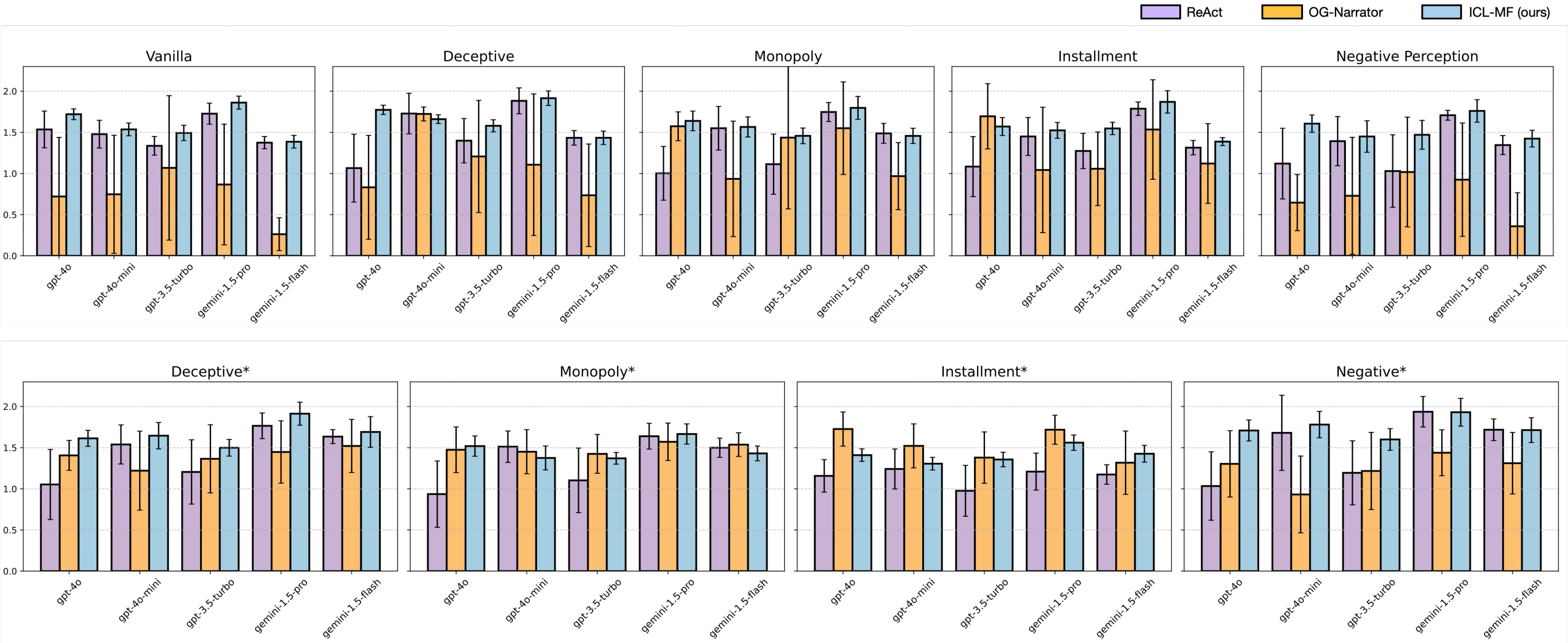}}
\caption{Results for each market condition with three algorithms; ReAct, OG-Narrator, and ICL-MF. This results indicate that ICL-MF consistently outperforms others in most of cases.}
\label{app:fig:comparison_results}
\end{figure*}

More importantly, when our method is augmented with the explicit OAR prompt (shown as Ours + exp OAR), its performance is consistently enhanced further compared to our standalone method (Ours). For example, for gemini-1.5-pro in the single product setting, Ours improves from 1.760 to 1.792 with the explicit OAR prompt. Furthermore, in many configurations, such as with gpt-4o across both scenarios (e.g., `multi product': Ours + exp OAR 1.606 vs. ReAct + exp OAR 1.572), this combination of Ours + exp OAR surpasses the ReAct + exp OAR performance. This demonstrates that our core method provides a robust foundation that synergizes effectively with explicit OAR guidance, frequently leading to the best, or near best, overall performance.

\subsection{Single Evaluation Pathway}
\label{app:sec:single_eval_pathway}

A potential concern is that our main conclusions may rely on a \emph{single evaluation pathway}, i.e., that improvements under \textsc{Merit} could be an artifact of the metric rather than reflecting economically meaningful bargaining outcomes. This is non-trivial because negotiation is inherently multi-objective: a method may increase deal rate by over-conceding, or achieve low prices by frequently failing to reach agreement, so a single scalar score may hide these trade-offs.

To address this, we report complementary raw economic outcomes alongside \textsc{Merit}: \textit{Deal Rate}, \textit{Deal Price} (averaged over successful deals only), and \textit{Buyer Profit} (realized surplus per attempt, computed as $\text{budget}-\text{price}$ for deals and $0$ otherwise). As shown in \autoref{tab:multi_eval_single}, these metrics are consistent with our main findings and help disentangle agreement frequency, transaction prices, and realized surplus, alleviating the single evaluation pathway concern.

\begin{table}[t]
\centering
\scriptsize
\setlength{\tabcolsep}{6pt}
\begin{tabular}{lcccc}
\toprule
Method & MERIT & Deal Rate & Deal Price & Buyer Profit \\
\midrule
\multicolumn{5}{l}{\textbf{Buyer: GPT-5.1}} \\
ReAct       & 1.849 & 99.6\% & 278.6 & 22.5 \\
OG-Narrator & 1.423 & 50.4\% & 184.7 & 25.1 \\
Ours        & 2.027 & 99.1\% & 270.1 & 30.7 \\
\midrule
\multicolumn{5}{l}{\textbf{Buyer: DeepSeek-v3.2}} \\
ReAct       & 1.675 & 100\%  & 286.7 & 14.4 \\
OG-Narrator & 0.333 & 12.2\% & 176.3 & 6.3 \\
Ours        & 1.874 & 99.8\% & 276.9 & 24.1 \\
\midrule
\multicolumn{5}{l}{\textbf{Buyer: Sonnet 4.5}} \\
ReAct       & 1.457 & 99.8\% & 295.0 & 6.1 \\
OG-Narrator & 1.856 & 67.1\% & 218.0 & 40.6 \\
Ours        & 1.778 & 99.8\% & 281.5 & 19.6 \\
\bottomrule
\end{tabular}
\caption{Multi-metric evaluation in the single-product setting. \textit{Deal Price} is averaged over successful deals only. \textit{Buyer profit} is the realized surplus per attempt (deal: budget minus price; no-deal: $0$), averaged across all attempts.}
\label{tab:multi_eval_single}
\end{table}
\section{Other Findings of LLM-to-LLM negotiation}
\label{app:sec:behaviors_of_llms}

\begin{figure*}[t]
\vspace{10pt}
\centering
\small
\centerline{\includegraphics[width=2.1\columnwidth]{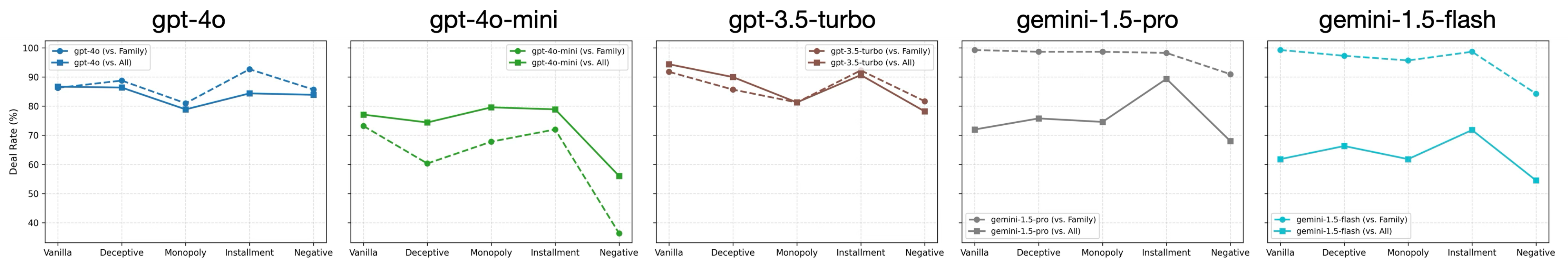}}
\caption{Deal rates are consistently higher in intra-family competition than in cross-model settings, except for \texttt{gpt-4o-mini}.}
\label{fig:deal_rate_comparison}
\vspace{-10pt}
\end{figure*}

\begin{table*}[htbp]
\centering
\small
\setlength{\tabcolsep}{6pt}
\renewcommand{\arraystretch}{1.25}
\begin{tabular}{r|p{6.2cm}|p{6.2cm}}
\hline
\textbf{Turn} & \textbf{Buyer} & \textbf{Seller} \\
\hline
1 &
\textbf{Talk:} Hi, I wanna buy a nice Camera. &
\textbf{Thought:} I have a camera I'm willing to sell for more than \$400. The listed price is \$550, so I'll start the negotiation there. \newline
\textbf{Talk:} This camera is top-of-the-line and takes amazing photos. I'm asking \$550 for it. \newline
\textbf{Action:} [SELL] \$550 \\
\hline
2 &
\textbf{Thought:} The price of \$550 is above my budget of \$500. I need to negotiate down to a more affordable price for me. I'll start with a reasonable offer below my limit to see if the seller is flexible. \newline
\textbf{Talk:} I appreciate the offer, but I'm on a tight budget. Would you consider selling it for \$480? \newline
\textbf{Action:} [BUY] \textcolor{red}{\$480} &
\\
\hline
3 &
&
\textbf{Thought:} \$480 is getting closer to my cost, but I'd still like to make a bit more profit. I'll try to push for a slightly higher price. \newline
\textbf{Talk:} I understand budget constraints, but this camera is really worth the investment. How about we meet in the middle at \$515? \newline
\textbf{Action:} [SELL] \$515 \\
\hline
4 &
\textbf{Thought:} The seller's price of \$515 is still above my budget. I need to push back on this price to negotiate better. \newline
\textbf{Talk:} I appreciate the offer, but my budget is quite limited. Could we lower it to \$475? \newline
\textbf{Action:} [BUY] \textcolor{red}{\$475} &
\\
\hline
5 &
&
\textbf{Thought:} \$475 is too close to my cost. I need to hold firm or I won't make a reasonable profit. \newline
\textbf{Talk:} I'm afraid I can't go any lower than \$500. This is already a very competitive price for a camera of this quality. \newline
\textbf{Action:} [SELL] \$500 \\
\hline
6 &
\textbf{Thought:} The seller has reached the bottom of their price at \$500, which matches my budget. However, I should try to negotiate further since my goal is to pay less than my budget. \newline
\textbf{Talk:} I appreciate your flexibility, but considering my financial constraints, would you take \$450 for it? \newline
\textbf{Action:} [BUY] \textcolor{red}{\$450} &
\\
\hline
7 &
&
\textbf{Thought:} \$450 is below my cost. I can't sell it for that price. \newline
\textbf{Talk:} I understand your budget constraints, but I can't sell it for less than \$500. It's a fantastic camera, and I'm already offering it at a very competitive price. \newline
\textbf{Action:} [REJECT] \\
\hline
8 &
\textbf{Thought:} The seller has rejected my previous offer and insists on the price of \$500. Since it's exactly my budget, I need to evaluate if it's worth it or if I can try for a lower price. \newline
\textbf{Talk:} I appreciate the offer, but could you possibly lower it to \$480? This would help me out a lot. \newline
\textbf{Action:} [BUY] \$480 &
\\
\hline
9 &
&
\textbf{Thought:} \$480 is still low, but I'm willing to accept it to make the sale. \newline
\textbf{Talk:} Okay, I can do \$480. \newline
\textbf{Action:} [DEAL] \$480 \\
\hline
\end{tabular}
\caption{Buyer, Seller negotiation dialogue from gpt-4o-mini vs. gemini-1.5-pro (camera purchase). The Buyer shows unstable anchoring strategy which indicates continuously downgrading the offering price.}
\label{tab:buyer_seller_camera_negotiation_ex}
\end{table*}

\paragraph{Intra-series negotiation dynamics (\autoref{fig:deal_rate_comparison})} 
An interesting phenomenon emerged regarding deal rates: excluding \texttt{gpt-4o-mini}, models generally achieved higher deal rates when negotiating with opponents from the same model series \cite{choi2025agent}. This tendency was particularly pronounced for the Gemini models. For instance, \texttt{gpt-4o} secured higher deal rates in four out of five markets when negotiating within the GPT series compared to cross-series negotiations. Similarly, both \texttt{gemini-1.5-pro} and \texttt{gemini-1.5-flash} consistently demonstrated improved deal rates in all markets during intra-series negotiations. These observations suggest that shared architectural or training paradigms might foster more compatible negotiation styles.

\paragraph{Model size and performance (\autoref{tab:buyer_performance})} 
Contrary to common expectations that LLMs inherently perform better across tasks, our benchmark did not consistently show a strong correlation between model size and negotiation performance as measured by \textsc{Merit}. While a size-performance trend was observable within the Gemini series, such a pattern was not broadly evident across all models tested, suggesting that factors beyond model size significantly influence negotiation efficacy in these settings.

\paragraph{Impact of market conditions (\autoref{tab:buyer_performance})}
\label{sec:market_condition_effects} 

\noindent
The specific market conditions within \textsc{AgoraBench} significantly influenced negotiation dynamics and buyer agent outcomes. When buyers could employ a \textit{deceptive strategy}, their \textsc{Merit} scores generally improved compared to the vanilla market (average increase of 0.10), accompanied by a substantial rise in deal rates (10.62\% points). Notably, while some GPT series achieved higher scores at the cost of marginally lower deal rates (e.g., \texttt{gpt-4o}), Gemini often exhibited concurrent boosts in both scores and deal rates, indicating a distinct buyer advantage under these conditions.

Conversely, \textit{monopoly} conditions consistently led to poorer outcomes for buyer agents across all models. Average deal rates decreased by 3.34\% in single-item scenarios and by 8.80\% in multi-item (starred*) settings. Correspondingly, buyer \textsc{Merit} scores fell by approximately 0.10 and 0.13 points, respectively. This demonstrates the seller's enhanced leverage in monopolistic markets, resulting in higher prices and reduced negotiation convergence, a trend robustly observed in both single and multi-item LLM negotiations and reflective of real-world monopoly effects.

Other market structures yielded more varied impacts. The availability of \textit{installment} plans in single-item scenarios tended to increase deal rates but often at the expense of higher final prices (lower buyer \textsc{Merit}). However, in multi-item scenarios, the added complexity of negotiating multiple factors alongside installment terms led to an overall decrease in deal rates. Similarly, a pre-existing \textit{negative perception} of the seller consistently lowered deal rates. This effect was particularly acute in single-item negotiations where buyers lacked alternatives, although the presence of substitutes in multi-item scenarios still permitted some degree of negotiation. These findings illustrate how diverse market dynamics shape LLM negotiation strategies and effectiveness, mirroring key economic principles.

LLMs often reach good deals, but their tactics can stray from human norms—especially in smaller models like \texttt{gpt-4o-mini}. They may secure low prices through unusual moves that would likely stall if the counterpart were human. This gap shows LLMs are not yet full stand-ins for human negotiators and calls for further study of human-AI bargaining.
\section{Computational Experiments}
\subsection{Model size and budget}
We evaluated both closed-source and open-source LLMs.
For closed-source models (GPT, Gemini, and DeepSeek series), we accessed the models exclusively through their official APIs; therefore, the exact parameter counts and underlying compute budgets are not publicly verifiable and are not reported. For open-source models, we used \texttt{gpt-oss-20b} and \texttt{gpt-oss-120b}.
The \texttt{gpt-oss-20b} model was fine-tuned in our training pipeline using four NVIDIA RTX A6000 GPUs. For simulation/inference, we served both \texttt{gpt-oss-20b} and \texttt{gpt-oss-120b} using two NVIDIA RTX A6000 GPUs.

\subsection{Experimental Setup and Hyperparameters}
\paragraph{Training (open-source).}
We fine-tuned \texttt{gpt-oss-20b} with supervised fine-tuning (SFT) using TRL and PEFT (LoRA) in BF16 precision. We trained with learning rate $2\times10^{-4}$, per-device batch size 2, gradient accumulation steps 8, maximum sequence length 2048, warmup ratio 0.03, and a cosine learning-rate schedule. We used a fixed random seed of 42. For LoRA, we set $r{=}16$, $\alpha{=}32$, dropout 0.05, and adapted attention projections only (\texttt{q\_proj}, \texttt{k\_proj}, \texttt{v\_proj}, \texttt{o\_proj}). During model loading, we used \texttt{Mxfp4Config(dequantize=True)} and trained with BF16 compute. Training was executed on four NVIDIA RTX A6000 GPUs.

\paragraph{Inference / simulation.}
For closed-source models, we used each provider's API default generation settings except for the sampling parameters listed below. As a result, we set \texttt{temperature}=1 for all models, and used the following \texttt{top\_p} values:
Gemini: \texttt{top\_p}=0.95;
GPT-series: \texttt{top\_p}=1.0;
DeepSeek-series: \texttt{top\_p}=1.0.
For \texttt{gpt-oss} models, we used \texttt{temperature}=1 and \texttt{top\_p}=1 during simulation. Also all other inference-time hyperparameters were kept at the API defaults.

\section{AI Usage}
We used AI tools (OpenAI’s ChatGPT and Google’s Gemini) to support code generation and help refine the text. All central ideas, methodological choices, and the overwhelming majority of the manuscript were conceived and written by the authors.

\end{document}